\documentclass[12pt,british,english,compact,a4paper]{article}
\usepackage[T1]{fontenc}
\usepackage[latin1]{inputenc}
\usepackage{bibentry}
\usepackage{pdflscape}
\usepackage{textcomp}
\usepackage{color}
\usepackage{tikz}
\usepackage{pgfgantt}
\usepackage{wrapfig}
\makeatother

\usepackage{babel}
\usepackage[top=2cm,bottom=2cm,left=2cm,right=2cm]{geometry}
\usepackage{color}
\usepackage{amsmath}
\usepackage{longtable}
\usepackage{amssymb}
\usepackage{graphicx}
 
\usepackage{blindtext}
\usepackage{hyperref}
\usepackage{sectsty}
\usepackage{enumitem}

\usepackage{multicol}
\usepackage{babel}
\usepackage[comma,sort&compress,numbers]{natbib}

\usepackage[small]{caption}
\usepackage[compact]{titlesec}
\usepackage{mdwlist}
\titlespacing{\section}{0pt}{*1}{*0.7}
\titlespacing{\subsection}{0pt}{*0.6}{*0.4}
\titlespacing{\subsubsection}{0pt}{*0.4}{*0.2}

\makeatletter

\renewcommand\thesubsection{\@arabic\c@subsection}
\sectionfont{\nohang\centering}
\makeatother

\usepackage{changepage}
\graphicspath{{figures/}}

\begin{document}

\begin{center}
\Large \textbf{A SIMULATION ENVIRONMENT FOR DRONE CINEMATOGRAPHY}\\

\vspace{10pt}
\normalsize

F. Zhang, D. Hall, T. Xu, S. Boyle and D. Bull \\
Bristol Vision Institute, University of Bristol, UK\\
\normalsize
\end{center}

\begin{adjustwidth}{50pt}{50pt}
\subsection*{ABSTRACT}

Simulations of drone camera platforms based on actual environments have been identified as being useful for shot planning, training and rehearsal for both single and multiple drone operations. This is particularly relevant for live events, where there is only one opportunity to get it right on the day. In this context, we present a workflow for the simulation of drone operations exploiting realistic background environments constructed within Unreal Engine 4 (UE4). Methods for environmental image capture, 3D reconstruction (photogrammetry) and the creation of foreground assets are presented along with a flexible and user-friendly simulation interface. Given the geographical location of the selected area and the camera parameters employed, the scanning strategy and its associated flight parameters are first determined for image capture. Source imagery can be extracted from virtual globe software or obtained through aerial photography of the scene (e.g. using drones). The latter case is clearly more time consuming but can provide enhanced detail, particularly where coverage of virtual globe software is limited.  The captured images are then used to generate 3D background environment models employing photogrammetry software. The reconstructed 3D models are then imported into the simulation interface as background environment assets together with appropriate foreground object models as a basis for shot planning and rehearsal. The tool supports both free-flight and parameterisable standard shot types along with programmable scenarios associated with foreground assets and event dynamics. It also supports the exporting of flight plans.  Camera shots can also be designed to provide suitable coverage of any landmarks which need to appear in-shot. This simulation tool will contribute to enhanced productivity, improved safety (awareness and mitigations for crowds and buildings), improved confidence of operators and directors and ultimately enhanced quality of viewer experience.
\end{adjustwidth}

\subsection*{INTRODUCTION}

Recently, unmanned aerial vehicles (UAVs) have been frequently employed as camera platforms in film and broadcast production to capture content offering an enhanced viewing experience. Drones offer flexible camera positioning with multiple angles and uninterrupted coverage, which are important for coverage of live events such as sports. In these cases, rehearsal opportunities are often limited and there is often only one opportunity for the directors, drone pilots and camera operators to conduct fly and shoot operations. A reliable and realistic simulation tool, supporting the integration of programmable foreground assets into realistic background environments, would therefore be of significant utility for planning, rehearsing and evaluating single and multiple drone operation in preparation for these types of event.  

There already exist commercial and royalty-free software packages capable of flight simulation within realistic \cite{nageli2017real} or virtual environments \cite{RealDrone,RealFlight}. Examples include DJI Flight Simulator \citep{DJIsimulat}, Google Earth Studio\cite{GoogleEarthStudio}, AirSim \cite{Airsim} and Microsoft Flight Simulator \cite{msflight}. DJI Flight Simulator focuses on training the flight skills of drone pilots, which offers a selection of virtual background scenarios and can simulate various weather conditions. However all its environmental models are not realistic. Google Earth Studio is a web-based animation tool based on Google Earth's satellite and 3D imagery, which can generate videos with an intuitive UI and features such as keyframe-based animation. Similar to Google Earth \cite{GoogleEarth} and Microsoft Bing Maps \cite{w:BingMap}, it provides sufficient resources for multiple view footage without providing any flight training or planning features. AirSim (Aerial Informatics and Robotics Simulation) is a plug-in package for Unreal Engine 4 \cite{w:UE4} and Unity \cite{w:Unity}. It has often been used as a platform for AI and control system research related to autonomous vehicles \cite{Shah2017AirSim}. However it is not furnished with shot type grammars and does not provide realistic environment assets. Microsoft Flight Simulator (2020) is currently under development for Windows 10 and Xbox One platforms, and is claimed to simulate the entire Earth using textures and topographical data from Bing Maps. It is designed primarily for flight training and simulation rather than drone cinematography, lacking features for shot type grammar and flexible foreground object integration. 

In this context, based on the preliminary work reported in \cite{c:Zhang28}, a new workflow for developing a fully functional CGI-based simulation tool with realistic background environments  is presented here. This also includes recommended parameters for the capture of 2D environmental images as a basis for 3D reconstruction. The prototype simulation interface supports flexible planning and training with example background and foreground 3D assets.

The rest of this paper is organised as follows. The proposed workflow for building a simulation tool with realistic background environments is firstly presented, and its primary steps, environmental image capture, 3D model reconstruction and simulation interface are further described in detail together with demonstration results and images. Finally, the conclusion is outlined alongside future research directions.

\subsection*{The Proposed Workflow}
\label{sec:workflow}

\begin{figure}[ht]
	\centerline{\includegraphics[width=1\linewidth]{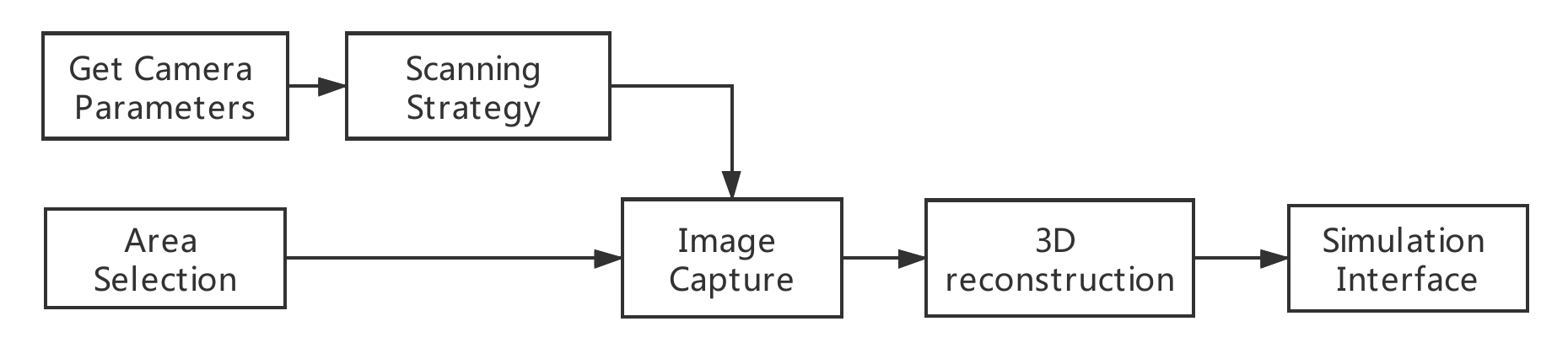}}
	\caption{Diagrammatic illustration of the proposed workflow.}
		\label{fig:workflow}
	\end{figure}

In order to simulate realistic environments and activities for drone cinematography, a workflow is proposed based on (i) environmental image capture, (ii) 3D reconstruction and (iii) a user-friendly simulation interface. This is depicted in Figure \ref{fig:workflow}. 

Given the location of the selected geographical area and the proposed camera parameters, the scanning strategy and its associated flight parameters are first determined for environmental image capture. The captured images are used to generate 3D background environment models employing photogrammetry techniques. The reconstructed 3D models are then imported into the simulation interface as background environment assets for further shot planning, training and rehearsal.

\subsection*{Environmental Image Capture}
\label{sec:images}

As an important step in photogrammetry, environmental image capture can be either conducted using a drone or helicopter camera platform at the actual geographical location or by photo scanning from specified viewing points within virtual globe software packages such as Google Earth \cite{GoogleEarth} and Microsoft Bing Maps \cite{w:BingMap}. In both approaches, the flight/scanning parameters influence the reconstruction quality of the 3D environment models. These parameters include flight/scanning trajectories, heights, viewing angles and picture overlaps ratios. Due to the limited computational capability, time and resource available, it is important to optimise these parameters to obtain as few images as possible but maintain the required reconstruction quality. 

\subsubsection*{Flight Trajectory}

\begin{figure}[htbp] 
	\centering    
	\includegraphics[width=0.8\linewidth]{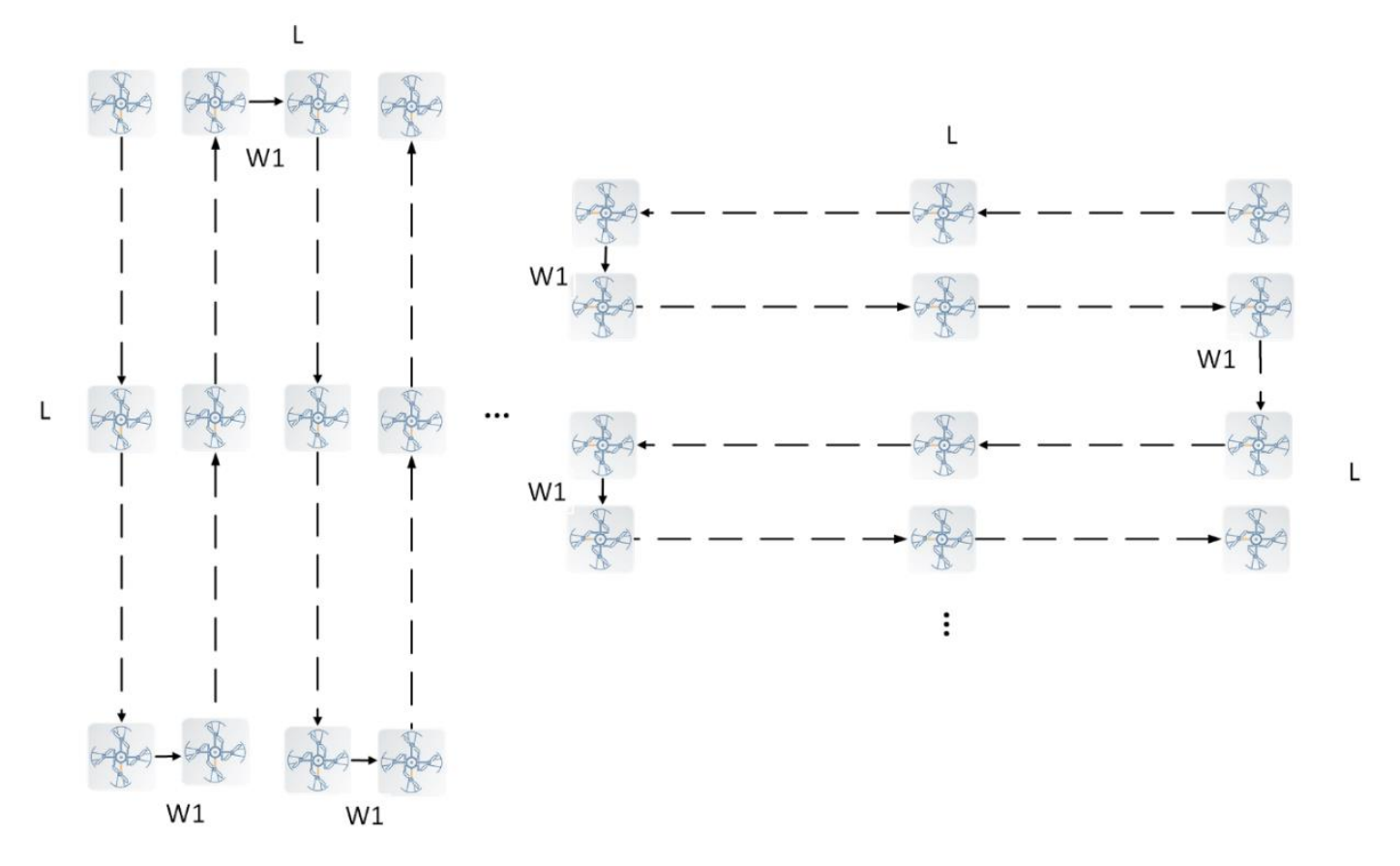}
	\caption{Grid scanning patterns. To cover of a area with a size of $L \times L$, the drone travels back and forth along straight lines at two orthogonal directions, with a cross-track distance of $W_1$.}
	\label{fig:flight trajectory}
	\end{figure}

The optimal flight trajectory for a specific background environment is highly dependent on the given landscape and object complexity \cite{smith2018aerial}. Notwithstanding this, the most commonly used flight pattern used in practise is grid scanning in two orthogonal horizontal directions. In order to simplify the scanning strategy for both shooting with real drones and capturing within virtual globe software packages, a grid scanning strategy has been employed in this work, as shown in Figure \ref{fig:flight trajectory}. 

\subsubsection*{Flight Heights}
\label{sec:heights}

The range of flight heights has been previously recommended in \cite{seifert2019influence}, where the results show that a single layer of scanning using a fixed height value can produce reasonably good reconstruction for a woodland landscape scenario. In this work, to generalise the height configuration for both landscape and urban environments, based on the recommendation in \cite{hawkins2016using}, we have employed a three-layer scanning approach. 

\begin{figure}[htbp]
\small
\begin{minipage}[b]{0.485\linewidth}
\centering
\centerline{\includegraphics[width=0.8\textwidth]{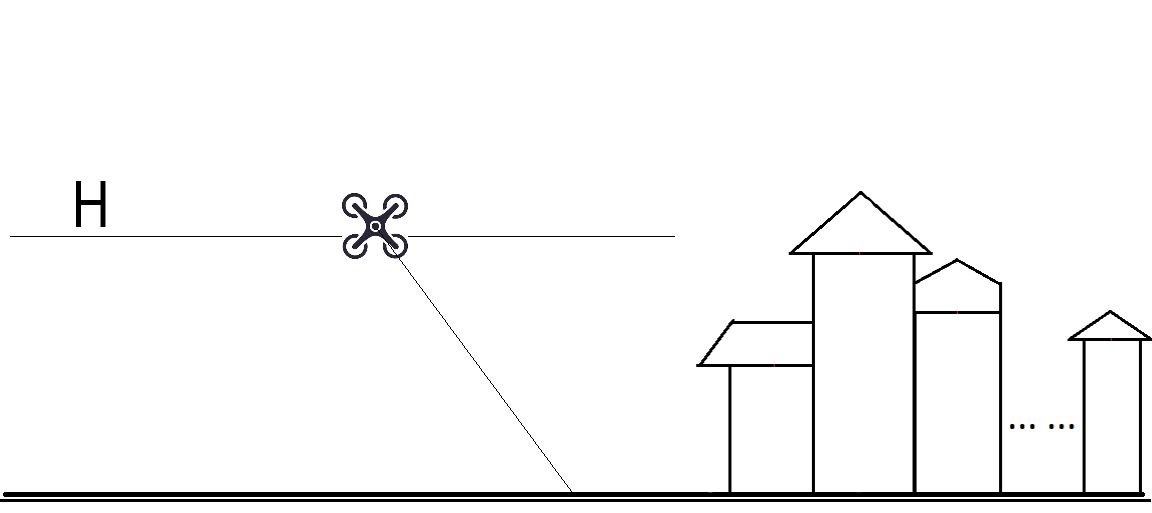}}
(a)Single height scanning in \cite{seifert2019influence}.
  
\end{minipage}
\begin{minipage}[b]{0.485\linewidth}
\centering
\centerline{\includegraphics[width=0.8\textwidth]{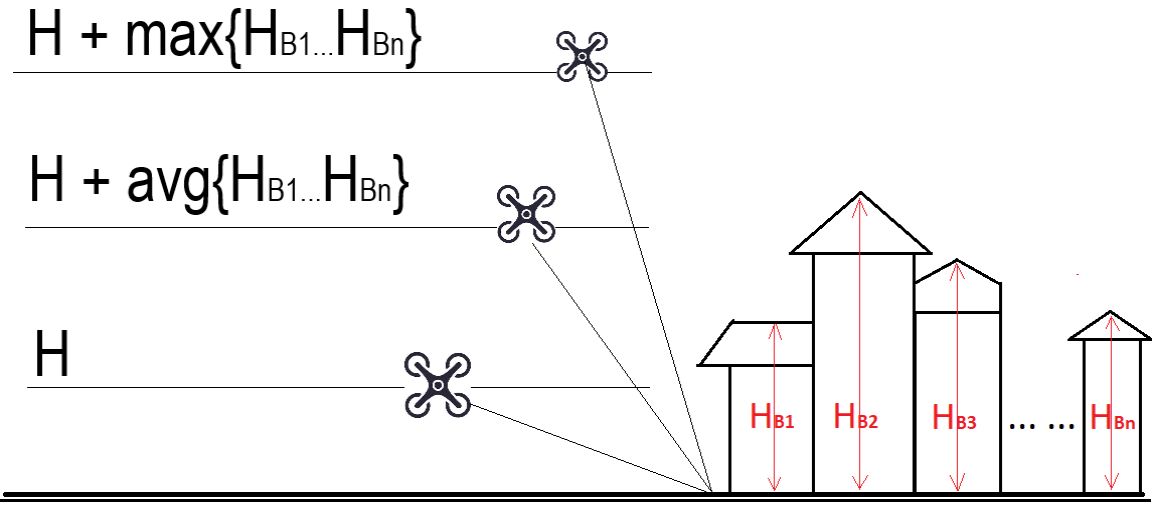}}
(b)Multiple height scanning proposed.

\end{minipage}
\caption{The flight height configuration}
\label{fig:Flight Height Reset}
\end{figure}

The first level, with a height of $H$, focuses on capturing most of the features of landscape. The second layer scanning (at $H + \mathrm{average~building~height}$) is used to obtain the details of objects below the average height in the covered area. The highest layer (at $H + \mathrm{maximum~building~height}$) is used to ensure everything (especially tall objects) can be properly covered during the scanning operation. This configuration is illustrated by Figure \ref{fig:Flight Height Reset} and equation (\ref{eq:heights}). 

\begin{equation}
\mathrm{Drone~Flight~Height} = \left\{
\begin{array}{ll}
H, &\mathrm{1^{st}~level};\\
H + \mathrm{average~building~height}, &\mathrm{2^{nd}~level};\\
H + \mathrm{maximum~building~height},&\mathrm{3^{rd}~level}.\\
\end{array}
\right.
\label{eq:heights}
\end{equation}
Here a fixed value of 20m is used for $H$ that is within the recommended range in \cite{seifert2019influence} for a default camera sensor size of 23.66$\times$13.3mm and a focal length 35mm. This should be adjusted based on the FOV (Field of View) relationship, given by (\ref{eq:fov}) and (\ref{eq:fov1}), for the actual camera settings used. 

\begin{equation}
\mathrm{FL} \times \mathrm{FOV} = \mathrm{SS} \times \mathrm{WD}
\label{eq:fov}
\end{equation}
Here FL and SS stand for focal length and sensor size of the camera respectively, and WD represents working distance (e.g. $H$ in this case). The actual working distance $\mathrm{WD_{act}}$ can be calculated by:
\begin{equation}
\mathrm{WD_{act}} = \frac{\mathrm{SS_{ref}}}{\mathrm{SS_{act}}} \times \frac{\mathrm{FL_{act}}}{\mathrm{FL_{ref}}} \times \mathrm{WD_{ref}}
\label{eq:fov1}
\end{equation}
in which $\mathrm{SS_{ref}}$ and $\mathrm{FL_{ref}}$ are default camera parameters as given above, while $\mathrm{WD_{ref}}$ is the recommended working distance (e.g. $H$=20m). $\mathrm{SS_{act}}$ and $\mathrm{FL_{act}}$ are the actual camera parameters used.  

\subsubsection*{Viewing Angles}
 
Viewing angle (the gimbal rotation angle on the drone) is another important shot parameter in photogrammetry scanning. In order to compare the reconstruction results for various viewing angles, three sets of angle parameters, including 90/67.5/45, 85/60/35, and 70/47.5/25 degrees, were employed, each of which has three different values for three height levels (from highest to lowest respectively) defined above. Other parameters such as flight trajectories, heights and picture overlap ratios are kept identical for three viewing angle sets. Here we used one of the high quality environmental assets \textit{Country Side} in UE4 market place \cite{countryside} as the source environment \cite{Lewis2012Simulating}. The captured images for each test set have been employed as inputs to a reconstruction software package, 3DF Zephyr \cite{3DFZephyr}, to generate a 3D environmental model.

\begin{figure}[htbp]
\small
\begin{minipage}[b]{0.325\linewidth}
\centering
\centerline{\includegraphics[width=1\linewidth]{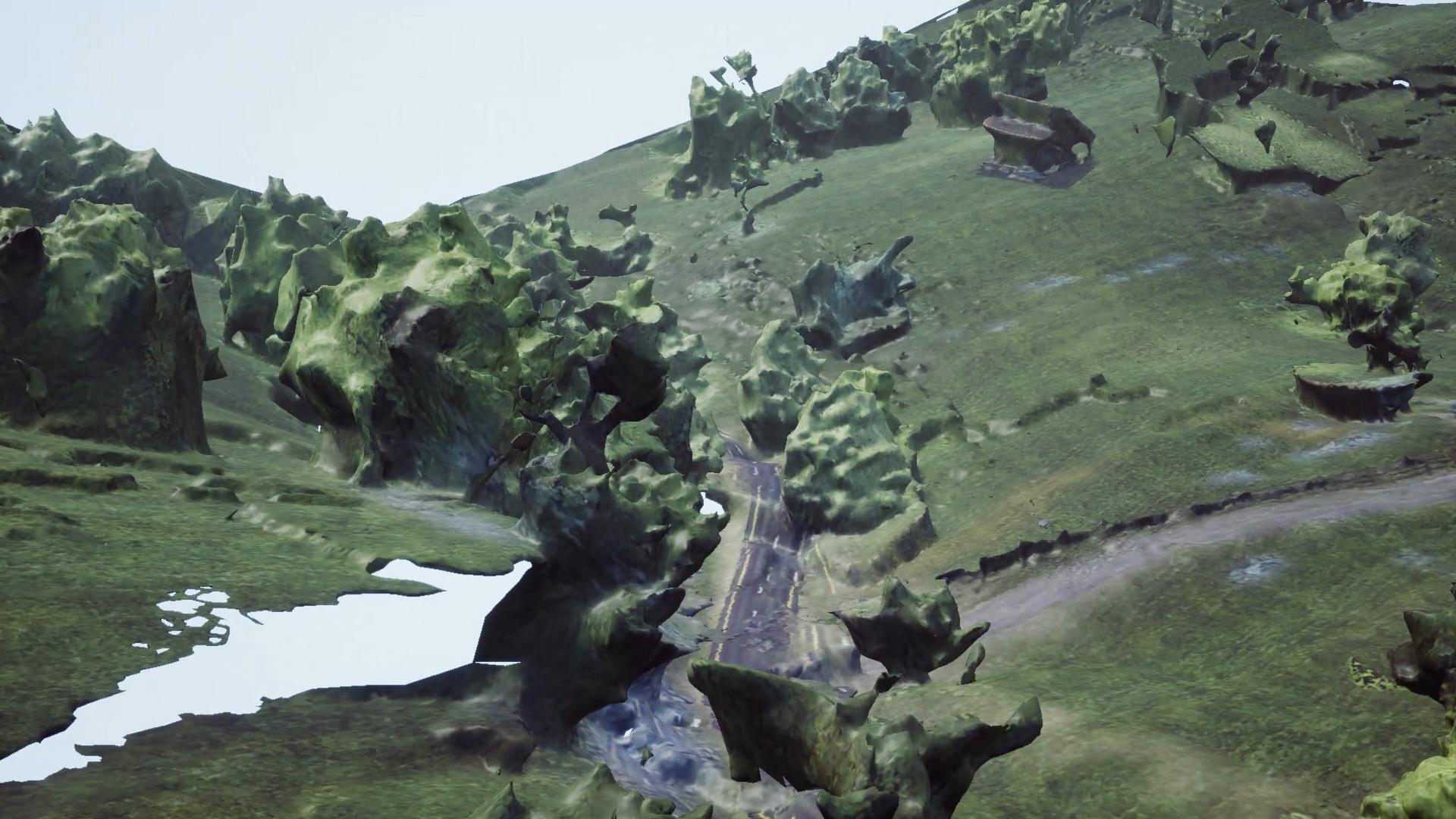}}
(a) High angles
\end{minipage}
\begin{minipage}[b]{0.325\linewidth}
\centering
\centerline{\includegraphics[width=1\linewidth]{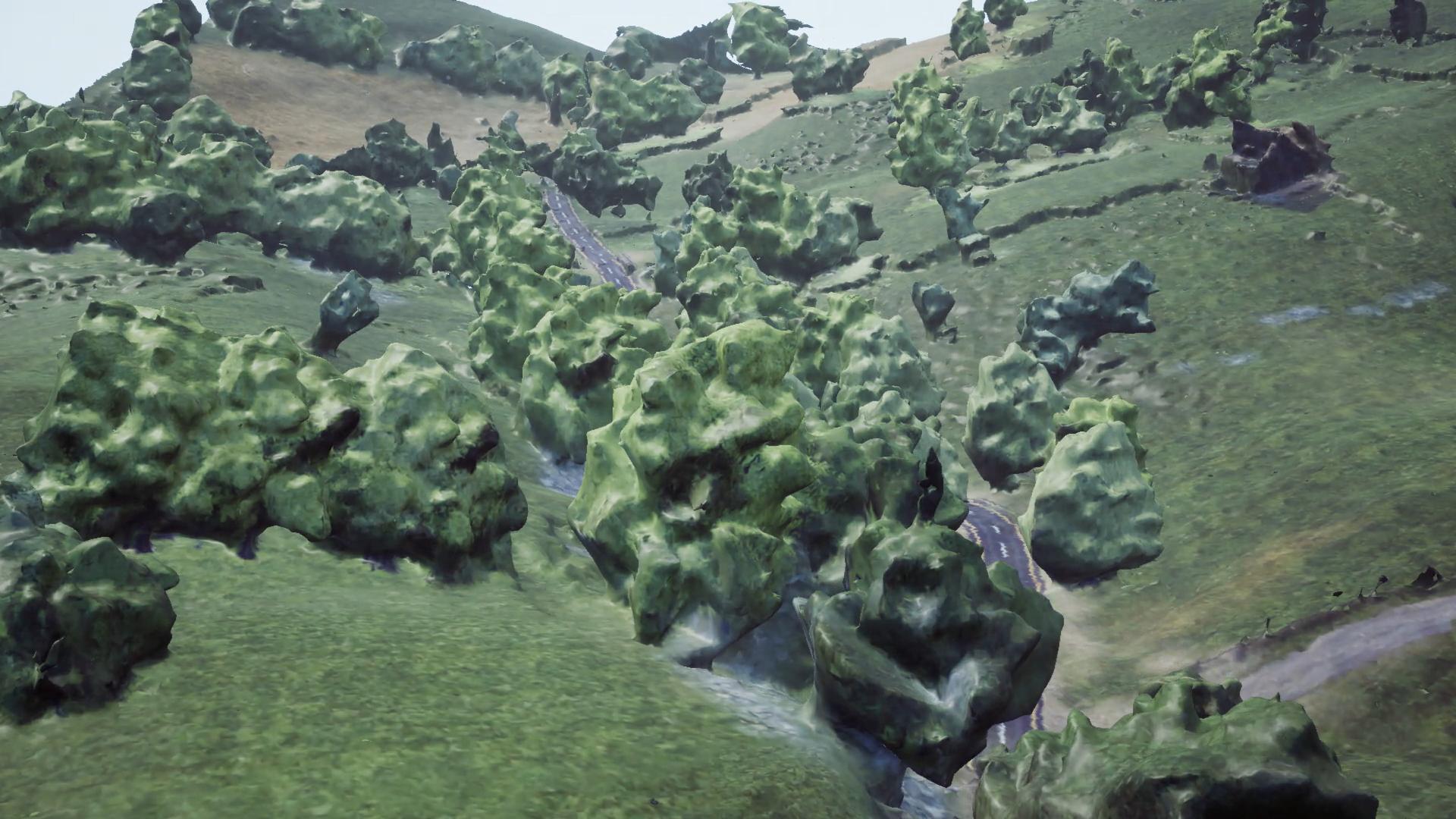}}
(b) Low angles
\end{minipage}
\begin{minipage}[b]{0.325\linewidth}
\centering
\centerline{\includegraphics[width=1\linewidth]{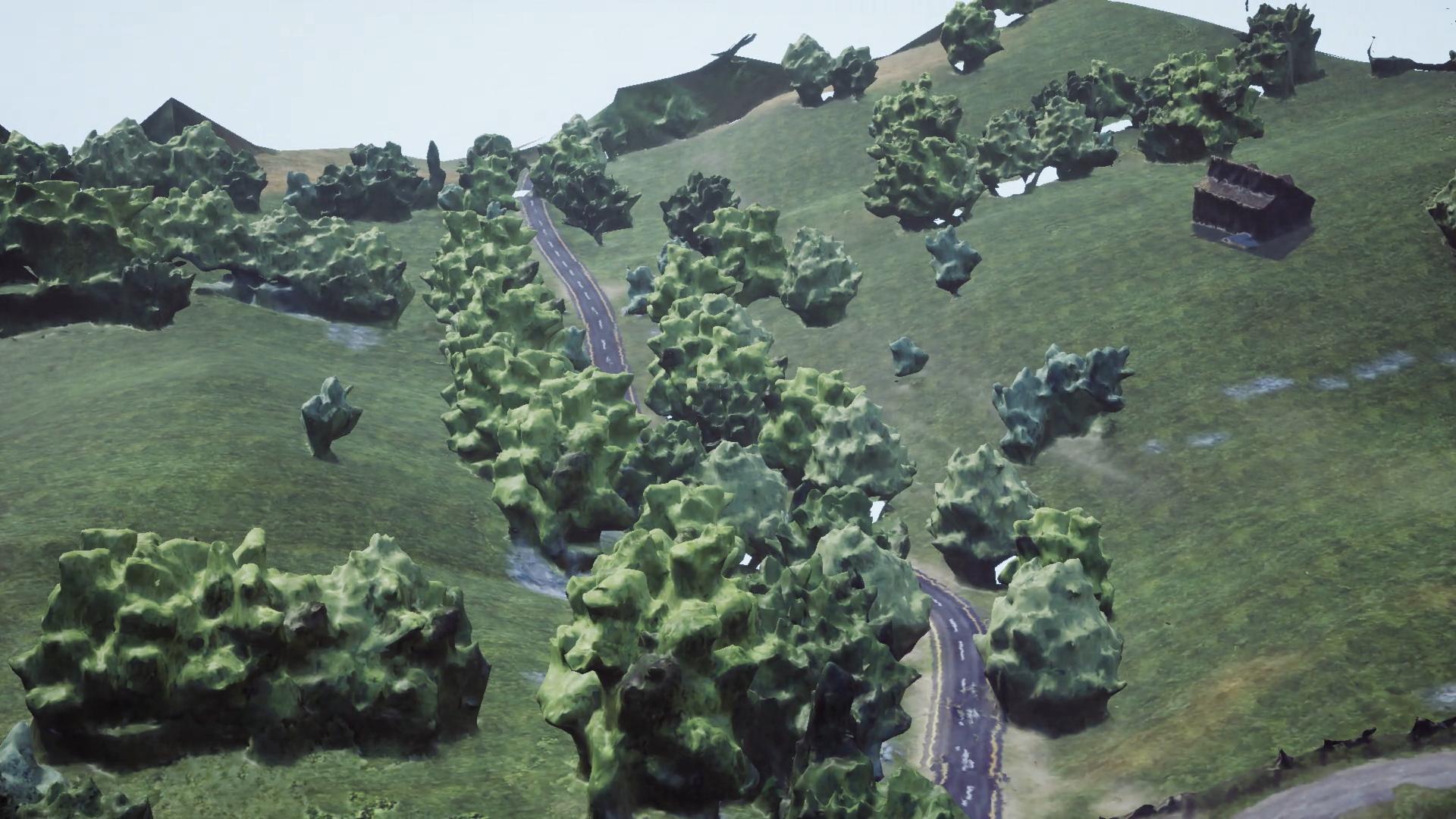}}
(c) Intermediate angles
\end{minipage}
\caption{Sample images of the reconstructed models based on images captured using three viewing angle sets.}
\label{fig:Viewing Angles}
\end{figure}

Example images of the reconstructed 3D models for three different viewing angle sets are shown in Figure \ref{fig:Viewing Angles}. It can be observed that high angles (90/67.5/45 degrees) lead to significant distortions of the landscape and the road, but provide good reconstruction for certain local details. The low angles (70/47.5/25 degrees) perform well on the shape of landscape but there are localised areas with high distortions. The intermediate angles (85/60/35 degrees) offer the best overall performance, with better reconstruction of landscape and less local detail loss. 

\subsubsection*{Overlapping Ratios}

\begin{figure}[htbp]
\small
\centering
	\begin{minipage}{0.325\linewidth}
  \centerline{\includegraphics[width=1\linewidth]{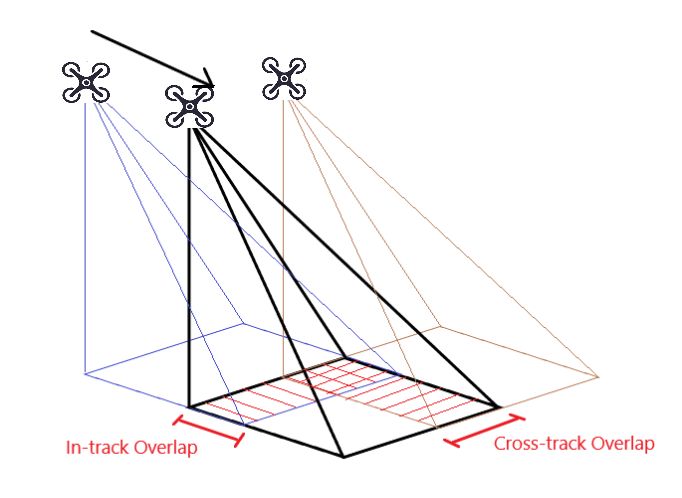}}
	(a)
\end{minipage}
	\begin{minipage}{0.325\linewidth}
  \centerline{\includegraphics[width=1.1\linewidth]{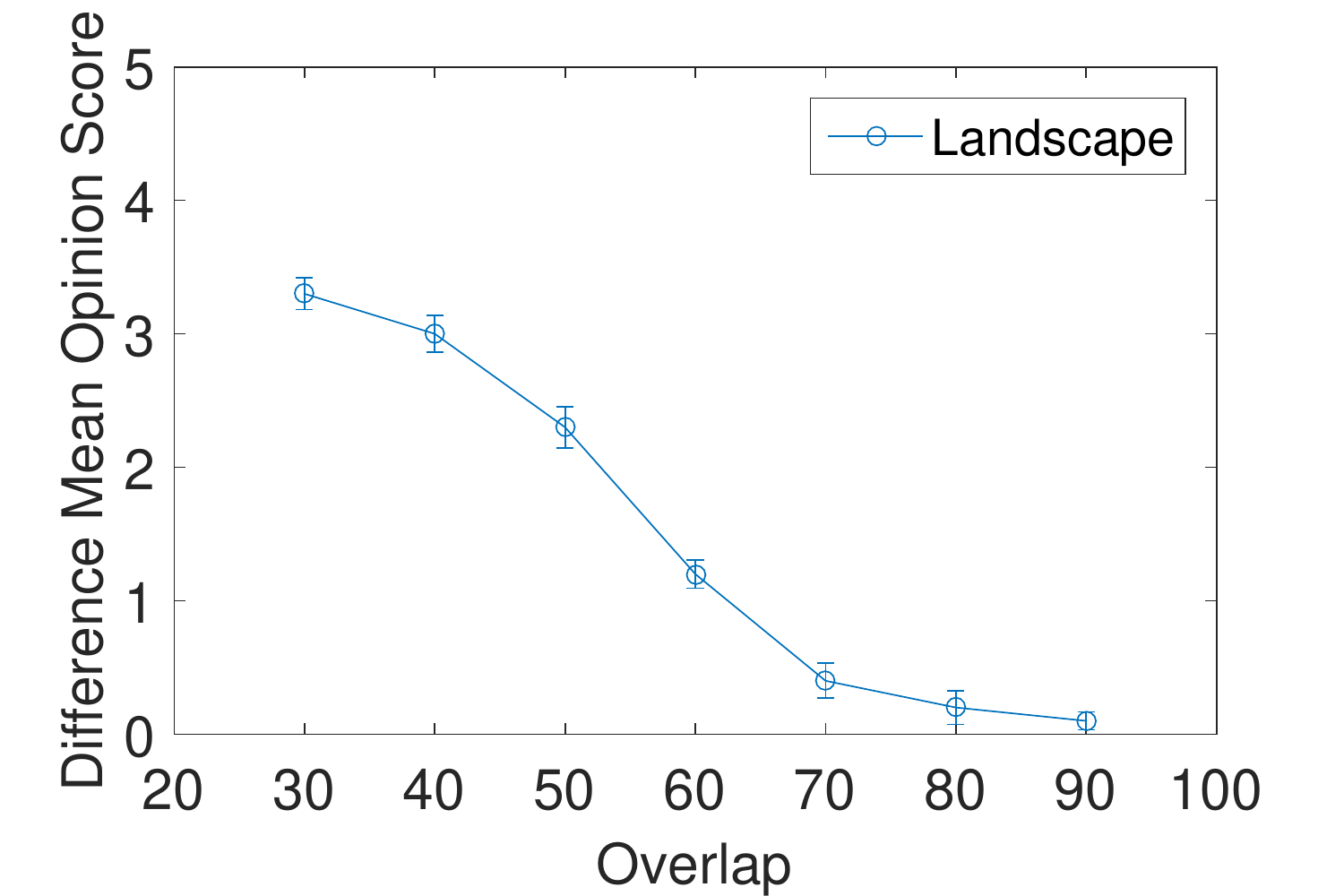}}
	(b)
\end{minipage}
\begin{minipage}{0.325\linewidth}
  \centerline{\includegraphics[width=1.1\linewidth]{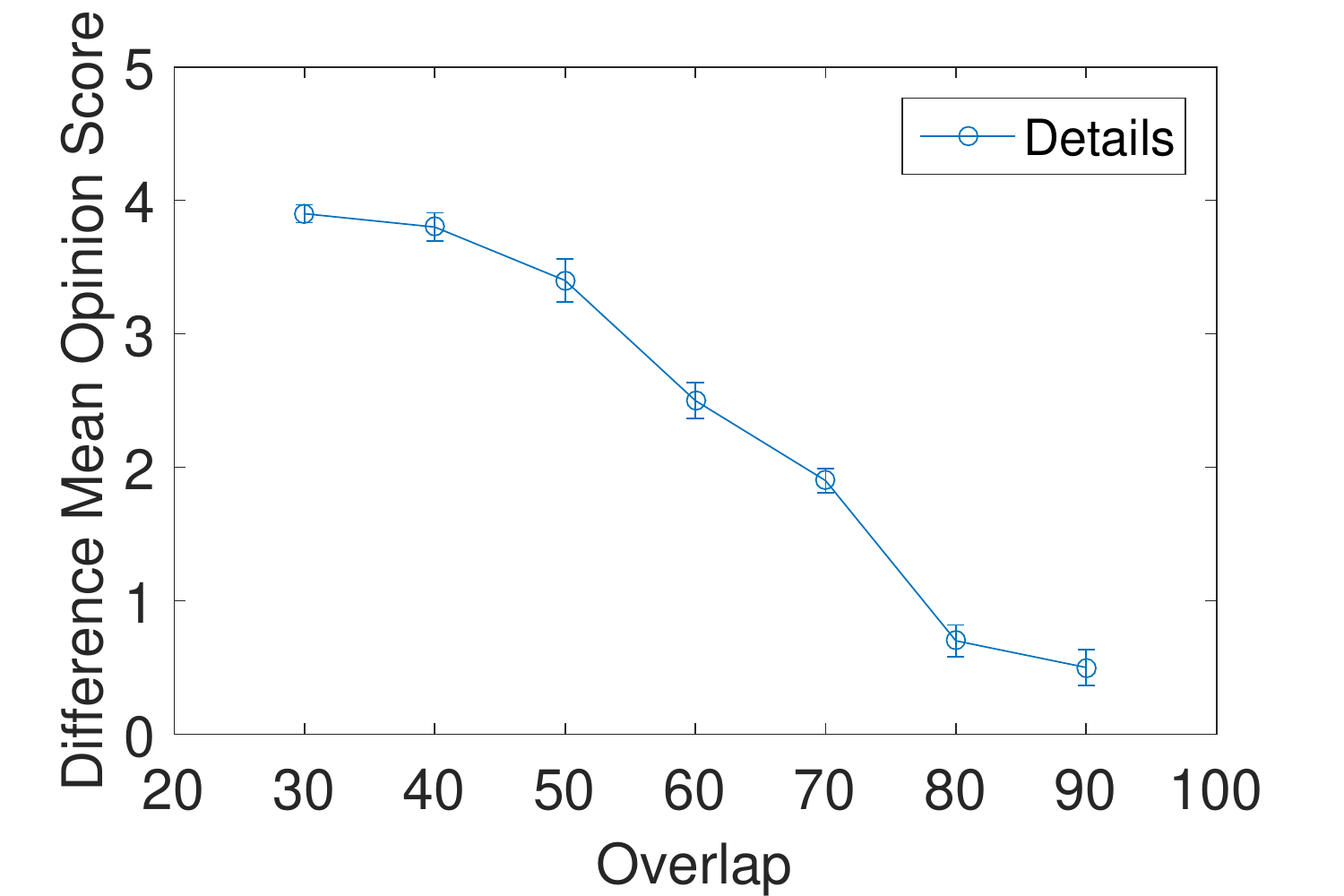}}
	(c)
\end{minipage}
	\caption{(a) Illustration of in-track and cross-track overlaps. (b,c) The DMOS scores collected for 3D models based on different in-track overlap ratios.}
  \label{fig:overlap}
\end{figure}

As shown in Figure \ref{fig:overlap}.(a), when environmental images are captured, the overlap between adjacent images (or video frames) captured in the same flight track is defined as in-track overlap, while the cross-track overlap is defined as the overlap between the adjacent images (or video frames) captured at neighbouring flight tracks. It is noted that these two overlap ratios are both related to the total number of images, which determine the reconstruction quality and efficiency.

\begin{figure}[ht]
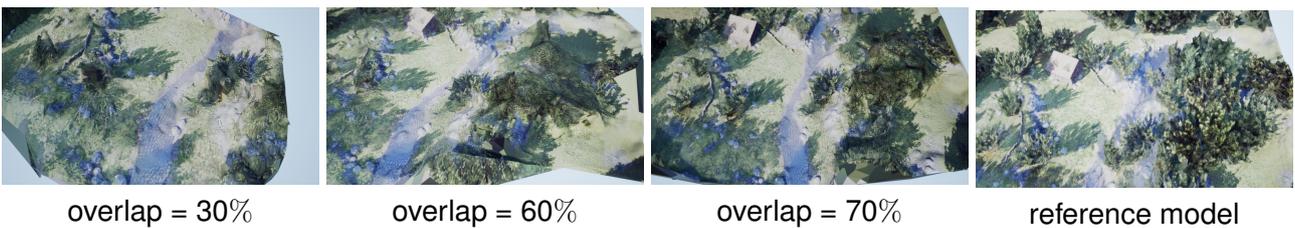

\small
\begin{minipage}{0.245\linewidth}
  \centerline{\includegraphics[width=\linewidth]{30g.pdf}}
  \centerline{overlap = 30$\%$}
  \label{landscape(30)}
\end{minipage}
\begin{minipage}{0.245\linewidth}
  \centerline{\includegraphics[width=\linewidth]{60g.pdf}}
  \centerline{overlap = 60$\%$}
  \label{landscape(60)}
\end{minipage}
\begin{minipage}{0.245\linewidth}
  \centerline{\includegraphics[width=\linewidth]{70g.pdf}}
  \centerline{overlap = 70$\%$}
  \label{landscape(70)}
\end{minipage}
\begin{minipage}{0.245\linewidth}
  \centerline{\includegraphics[width=\linewidth]{ref.pdf}}
  \centerline{reference model}
  \label{landscape(95)}
\end{minipage}
\caption{Example images showing 3D models based on different in-track overlap ratios (for landscape).}
\label{fig:landscape}
\end{figure}

\begin{figure}[ht]
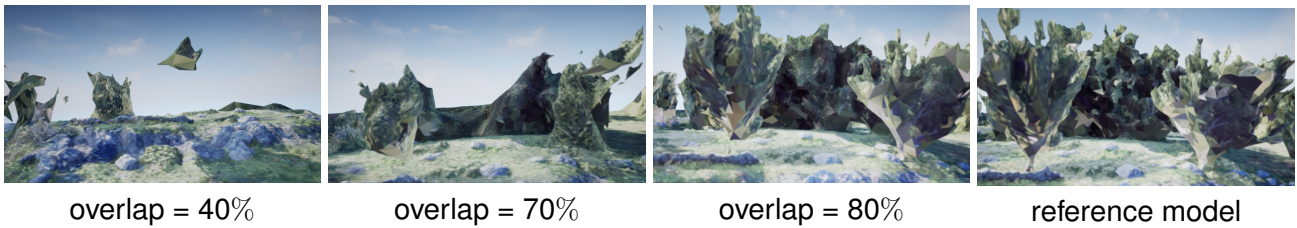

\small
\begin{minipage}{0.245\linewidth}
  \centerline{\includegraphics[width=\linewidth]{40d.pdf}}
  \centerline{overlap = 40$\%$}
  \label{details(40)}
\end{minipage}
\begin{minipage}{0.245\linewidth}
  \centerline{\includegraphics[width=\linewidth]{70d.pdf}}
  \centerline{overlap = 70$\%$}
  \label{details(70)}
\end{minipage}
\begin{minipage}{0.245\linewidth}
  \centerline{\includegraphics[width=\linewidth]{85d.pdf}}
  \centerline{overlap = 80$\%$}
  \label{details(80)}
\end{minipage}
\begin{minipage}{0.245\linewidth}
  \centerline{\includegraphics[width=\linewidth]{refd.pdf}}
  \centerline{reference model}
  \label{details(95)}
\end{minipage}
\caption{Example images showing 3D models based on different in-track overlap ratios (for details).}
\label{fig:details}
\end{figure}

The selection of both overlap ratios has been previously reported \cite{ventura2016low,james2012straightforward,hawkins2016using}, with recommendations between 60\% and 75\% for practical use. To further investigate their influence on 3D reconstruction quality, seven different in-track overlap values (30\%, 40\%, 50\%, 60\%, 70\%, 80\% and 90\%) were employed to capture 2D images for a selected small area (50m$\times$50m) within the \textit{Country Side} environmental asset from the UE4 marketplace \cite{countryside}. Here we used the same flight height and trajectories, viewing angles and cross-track overlap\footnote{We fixed cross-track overlap ratio here to solely test in-track overlap. Actually the experimental results for in-track overlap will be still valid for cross-track overlap if image size and aspect ratio are taken into account.} (70\%).

Based on each reconstructed model (and within the original 3D asset), a ten second free flying shot was generated using UE4, with identical flight path. All eight video clips (seven test videos plus an original) were viewed by 15 participants using a double stimulus continuous quality scale (DSCQS) methodology \cite{r:BT500}. All the participants were requested to provide subjective scores for both landscape shapes and local details.

The collected DMOS (Differential Mean Opinion Score) results, comparing the reference video (ground truth) and each test version for both landscape and details, are plotted in Figure \ref{fig:overlap}.(b,c). It can be observed that, as the in-track overlap ratio increases, the DMOS values becomes lower for both landscape and details. To achieve relatively high reconstruction quality with DMOS being lower than 1, the overlap ratio needs to be greater than 70\% for landscape and 80\% for details. This characteristics can also be observed in Figure \ref{fig:landscape} and \ref{fig:details}, where example figures of the reconstruction models using different in-track overlap ratios are illustrated. 

\subsection*{3D Model Reconstruction}
\label{sec:reconstruction}

\subsubsection*{Pre-processing}

The input image dataset obtained may contain artefacts due to photogrammetry errors or object motion, and these can result in significant distortions during reconstruction. Such defects can be removed or corrected through texture in-painting \cite{tschumperle2005vector,wong2008nonlocal}. It is noted that most in-painting algorithms are relatively complex and time consuming when processing a large number of images. During the generation of our demonstration results, in order to achieve efficient reconstruction, simple manual outlier rejection was applied instead on the input image dataset.

\subsubsection*{Photogrammetry Reconstruction}

Numerous open source and commercial photogrammetry software packages are available for producing 3D models from multiple view 2D images. Notable examples include Autodesk ReCap \cite{AutoDeskRecap}, 3DF Zephyr \cite{3DFZephyr}, and Pix4D Mapper \cite{pix4d}. Autodesk Recap produces relatively poor results from Google Earth captured images, and requires `cloud credits' to perform online analysis on AutoDesk servers. 3DF Zephyr Aerial generates 3D models with improved quality, especially for object-based scenarios \cite{c:Zhang28}. However it requires excellent local graphical calculation capability and large GPU memory for processing. Pix4D Mapper has been specifically designed for professional drone mapping. It creates reconstructions, for the same input images, with equivalent or better quality than 3DF Zephyr and with lower computation complexity. In this work, Pix4D mapper has therefore been adopted as the reconstruction software of choice. 


\subsubsection*{Post Processing}

\begin{figure}[ht]
  \centering
  \includegraphics[width=0.5\linewidth]{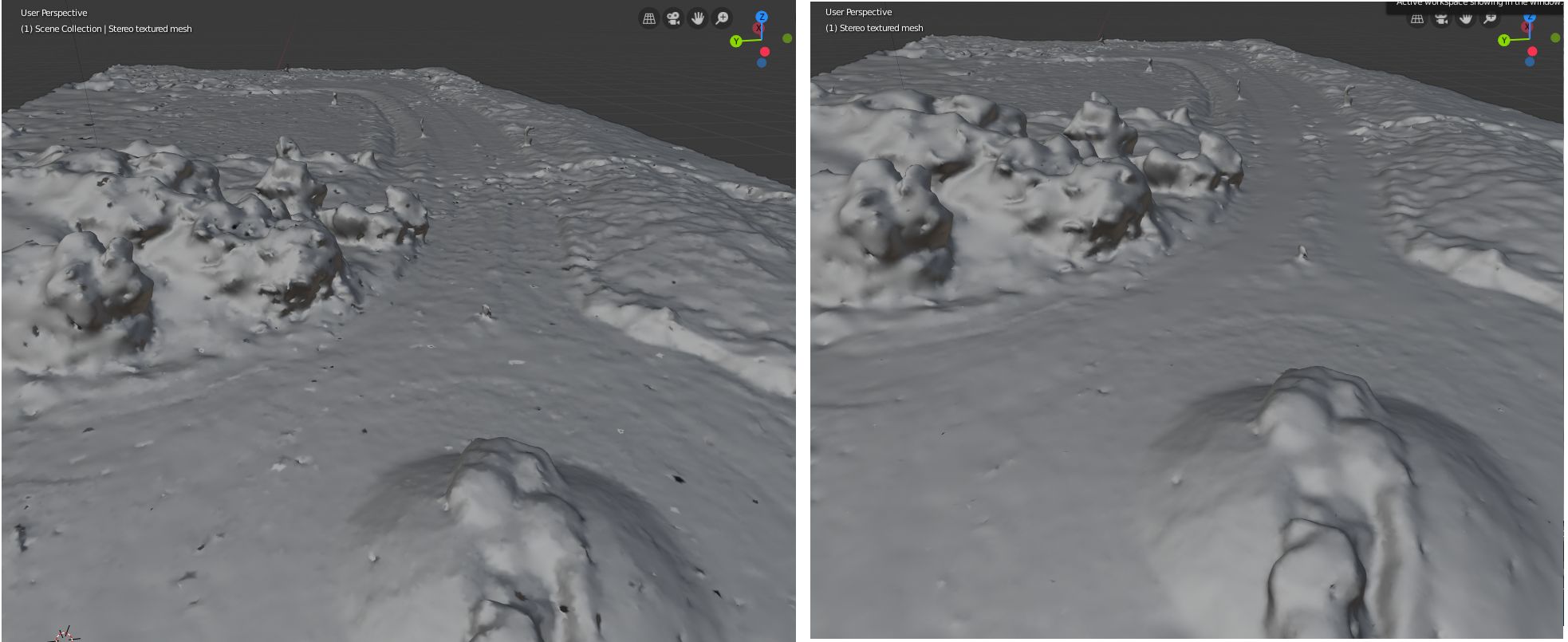}
  \caption{An example of using the Smooth tool in Blender for post-processing. (Left) The initial 3D model structure with visible bumpy artefacts on the flat surface. (Right) The processed 3D model structure after applying the Smooth tool.}
  \label{fig:smooth1}
\end{figure}

\begin{figure}[ht]
  \centering
  \includegraphics[width=0.5\linewidth]{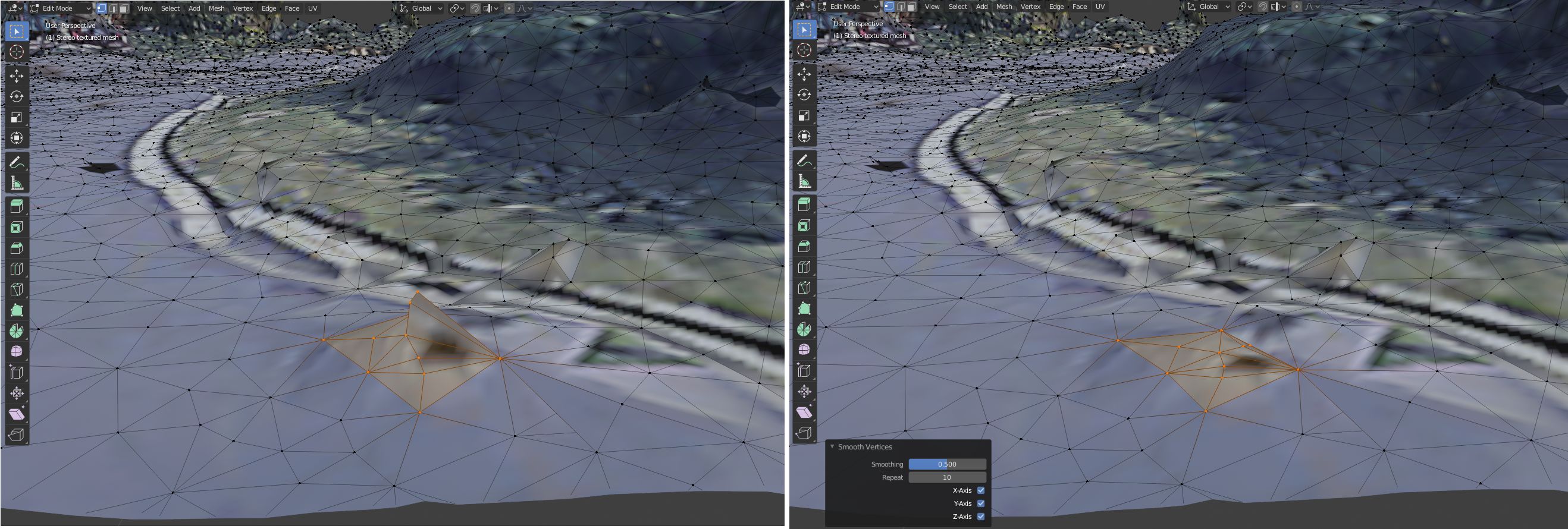}
  \caption{An example of using the Flatten tool in Blender for post-processing. (Left) 3D Model before Flatten vertices operation. (Right) 3D Model after Flatten vertices operation.}
  \label{fig:smooth3}
\end{figure}

\begin{figure}[ht]
  \centering
	\begin{minipage}{0.5\linewidth}
  \centerline{\includegraphics[width=\linewidth]{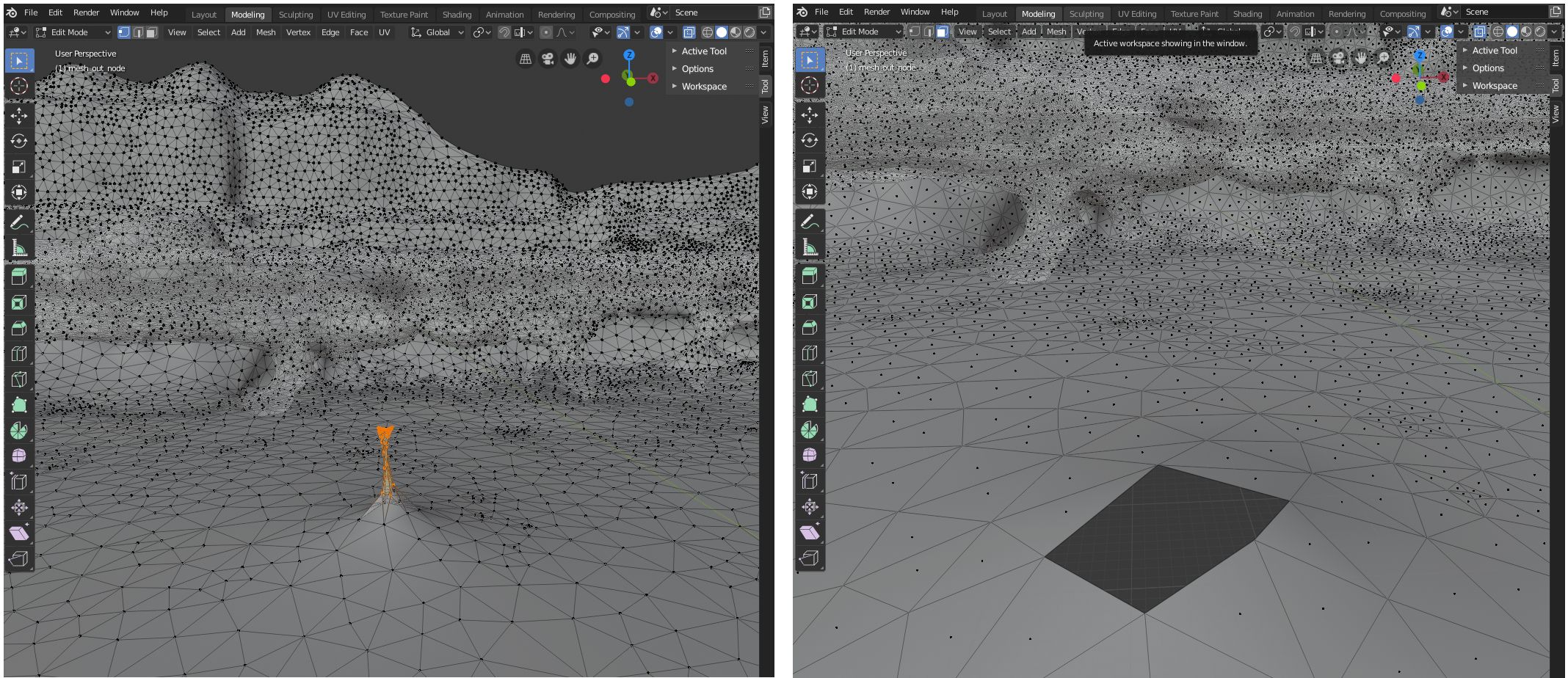}}
  \centerline{(a)  \ \  \ \ \ \ \ \ \ \ \ \ \ \ \ \ \ \ \ \   \ \ \ \ \ (b)}
\end{minipage}
\hfill
\begin{minipage}{0.5\linewidth}
  \centerline{\includegraphics[width=\linewidth]{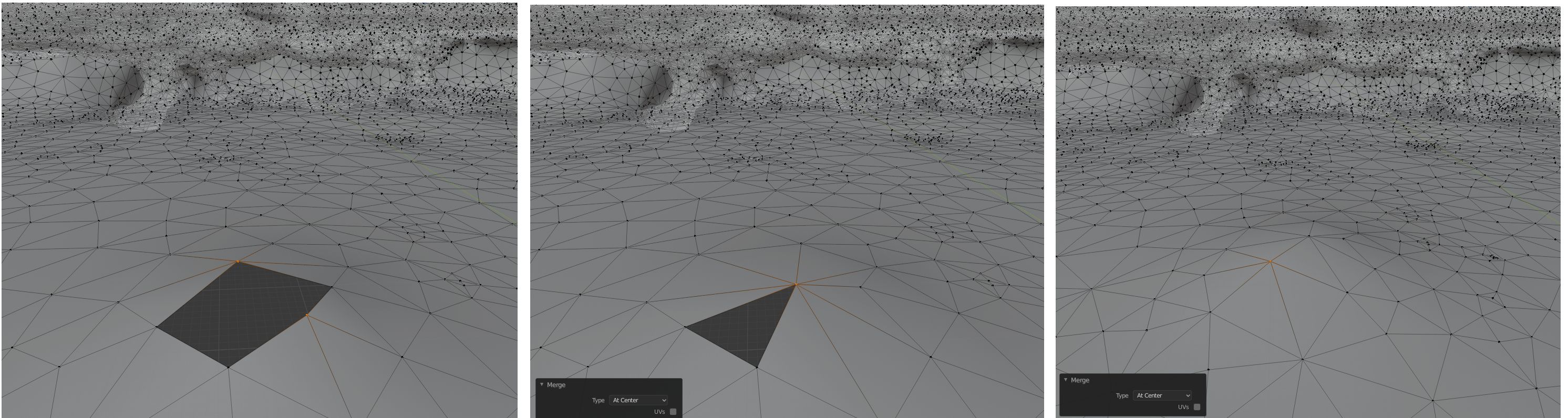}}
  \centerline{(c) \ \ \ \ \ \ \ \ \  \ \ \ \ \ (d) \ \ \ \ \ \ \ \ \ \ \ \ \  (e)}
\end{minipage}
\caption{An example of mesh modification using Blender. (a) A water surface that has been smoothed but still contains spikes where the mesh is highly distorted. (b) The artefact is deleted. (c-e) It is then filled with the similar textures as in the neighbouring area by applying Merge vertices operation.}
\label{fig:mesh}
\end{figure}

\begin{figure}[ht]
  \centerline{\includegraphics[width=0.5\linewidth]{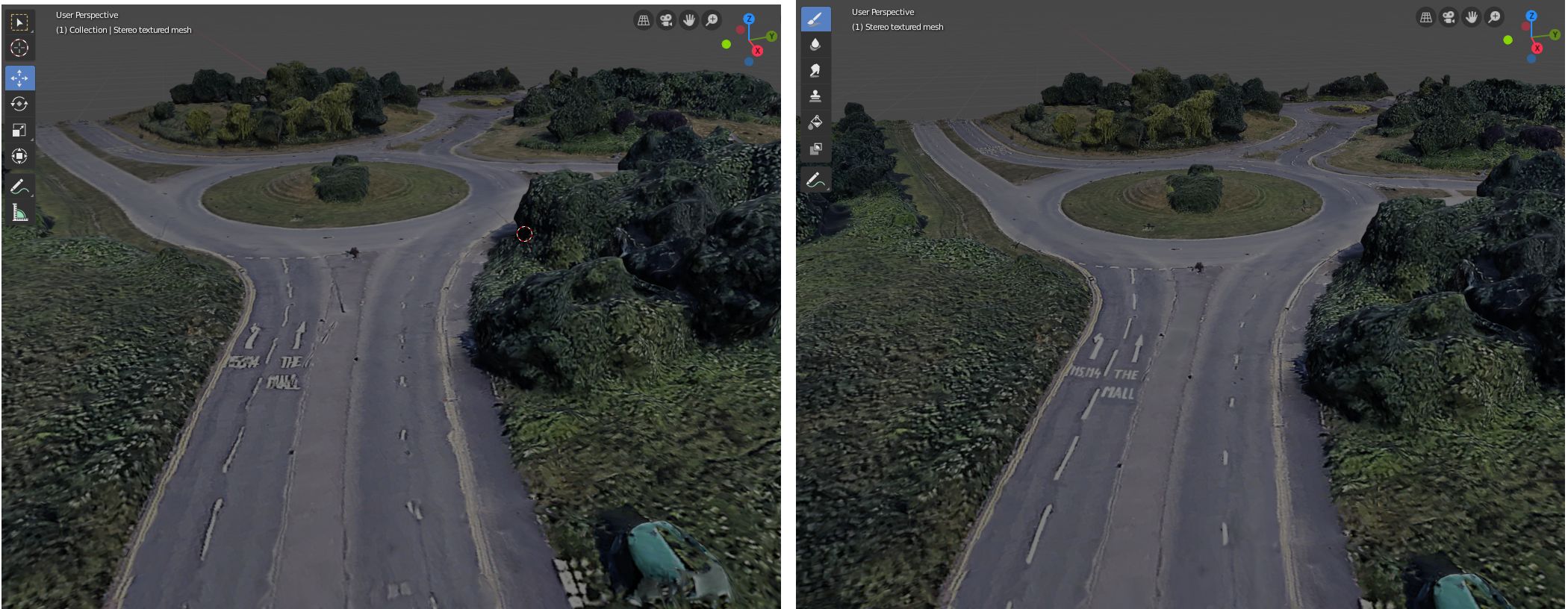}}
	\caption{(Left) A screen shot of the 3D Model before applying Texture Paint with distorted road markings. (Right) A screen shot of the 3D Model after applying Texture Paint with new markings re-painted in the same location.}
  \label{fig:blendtexedit}
\end{figure}

Although photogrammetry software can provide reasonably good reconstruction results, a large number of visible artefacts (e.g. bumps and holes) remain and these can impair viewing experience. 3D model editing (e.g. based on Blender \cite{blender}) can be employed to further correct these distortions. When an initial reconstructed model is imported into Blender, it can be used to enhance the 3D texture mesh, using features such as Surface Smoothing, Flatten Mesh Modification and Texture Painting. Example results are shown in Figure \ref{fig:smooth1}-\ref{fig:blendtexedit}.




Initial texture meshes may also contain areas of high distortion and irregularity. In cases when Smooth does not work, manual rebuilding is required. Figure \ref{fig:mesh}.(a) shows a water surface that has been smoothed but still contains spikes where the mesh is highly distorted. The artefact can be deleted (Figure \ref{fig:mesh}.(b)) and filled with the similar textures as in the neighbouring area by applying Merge vertices operation (Figure \ref{fig:mesh}.(c-e)).

%

\subsection*{The Simulation Interface}
\label{sec:interface}

The simulation interface was built using Unreal Engine 4 \cite{w:UE4} and has been packaged as a standalone application. This software provides three primary modes: editing, simulation and free play.

\subsubsection*{Pre-generated 3D Environments and Objects}

\begin{figure}[htbp]
\small
	\begin{minipage}{0.485\linewidth}
	  \centering
  \centerline{\includegraphics[width=\linewidth]{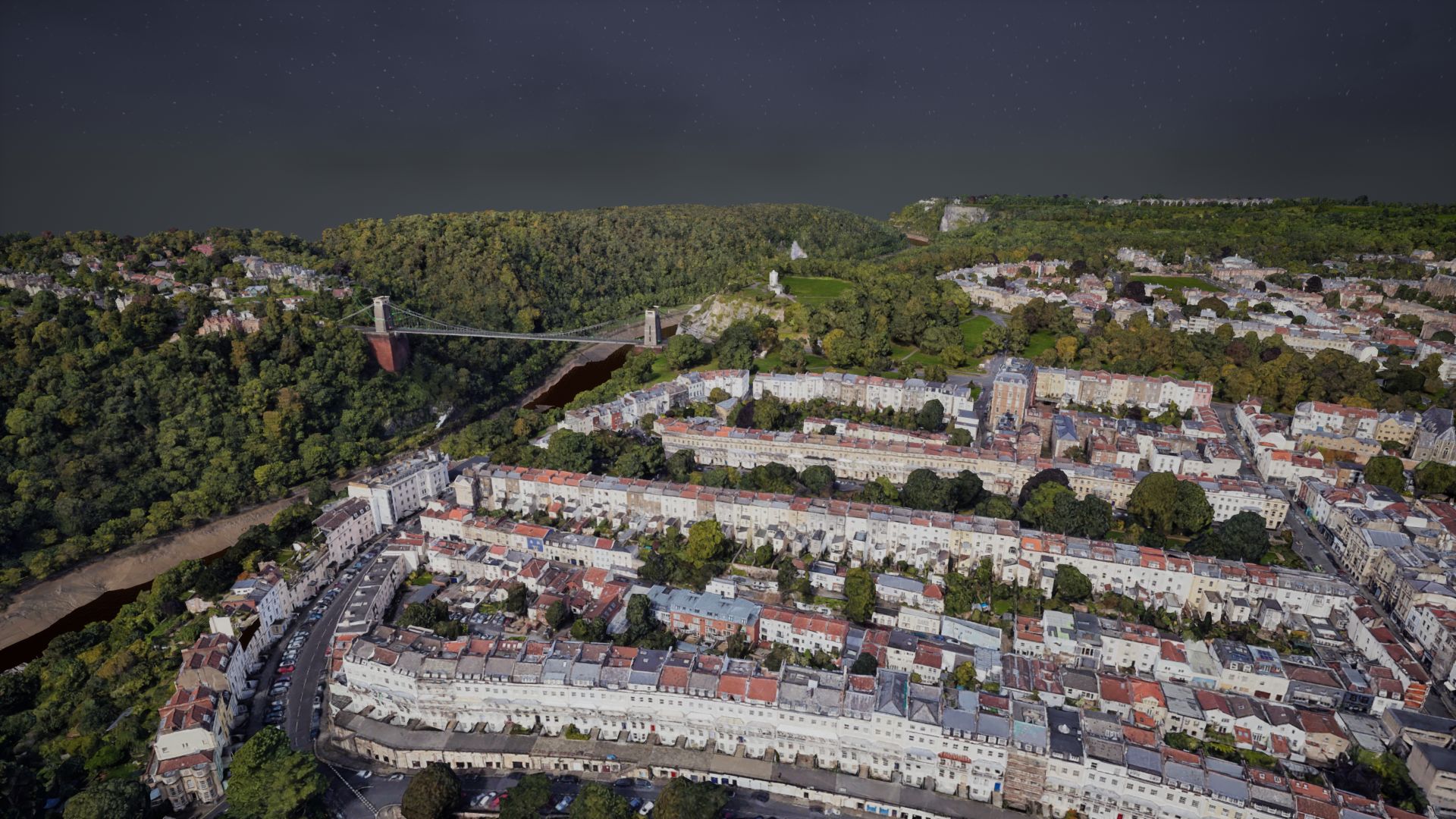}}
{Clifton, Bristol}
\end{minipage}
	\begin{minipage}{0.485\linewidth}
	  \centering
  \centerline{\includegraphics[width=\linewidth]{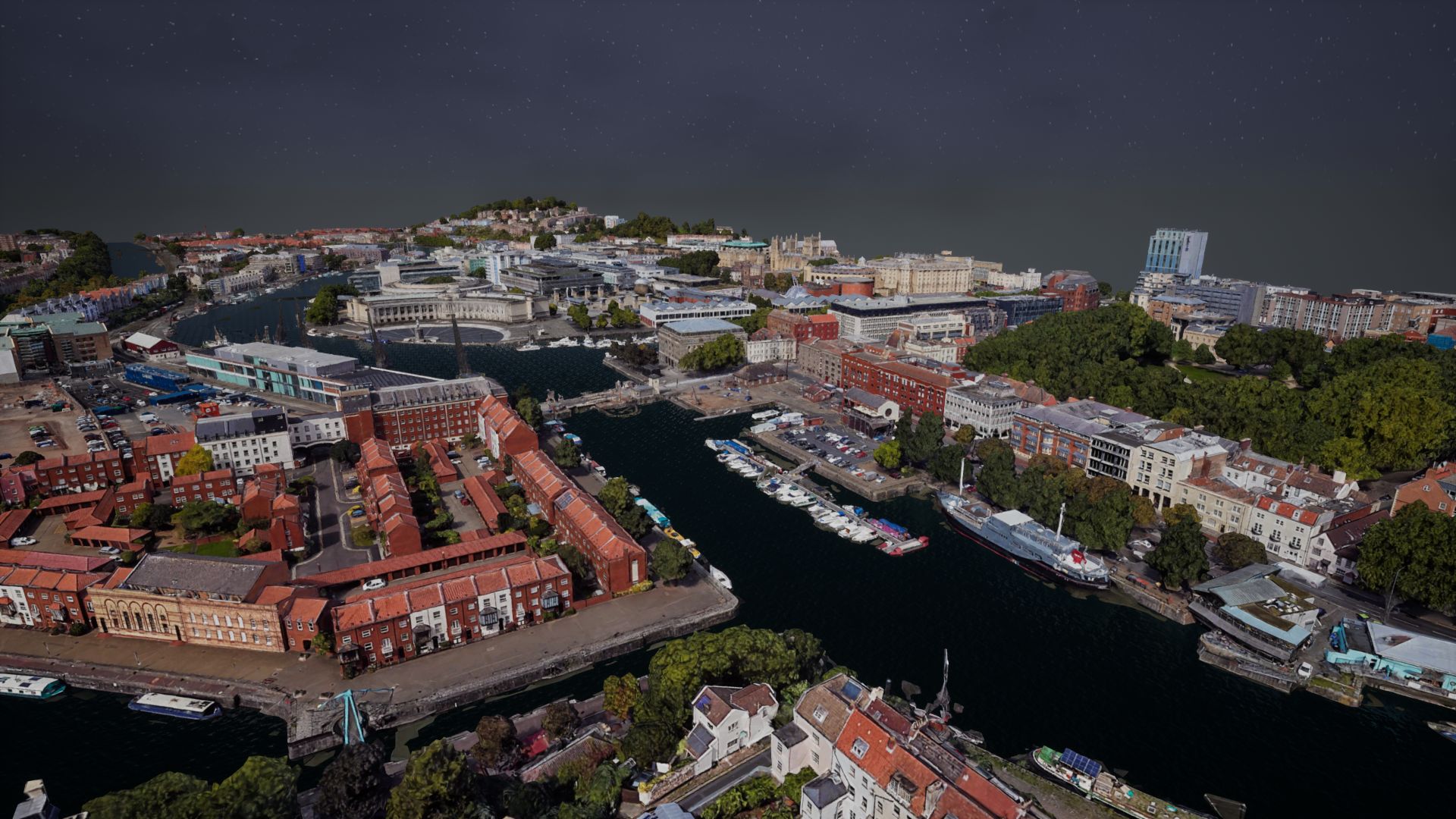}}
  {Harbour, Bristol}
\end{minipage}
	\caption{Pre-defined environments.}
	  \label{fig:models}
\end{figure}

Two example environments, Clifton Downs and Harbourside (both are within the city of Bristol, UK), have been pre-generated for this simulation tool using the workflow described above. Both of these are reconstructed based on the source data from Google Earth. Example images are shown in Figure \ref{fig:models}. 

\begin{figure}[ht]
  \centering
	\begin{minipage}{0.245\linewidth}
	  \centering
  \centerline{\includegraphics[width=\linewidth]{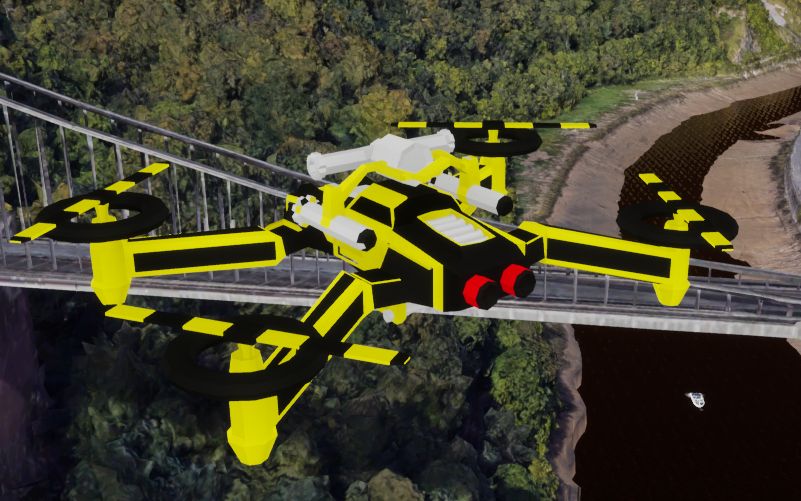}}
	(a)
\end{minipage}
	\begin{minipage}{0.245\linewidth}
	  \centering
  \centerline{\includegraphics[width=\linewidth]{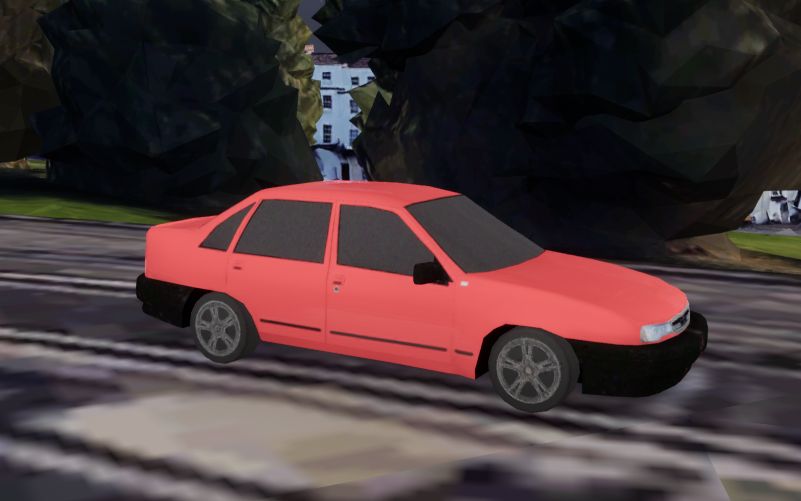}}
	(b)
\end{minipage}
	\begin{minipage}{0.245\linewidth}
	  \centering
  \centerline{\includegraphics[width=\linewidth]{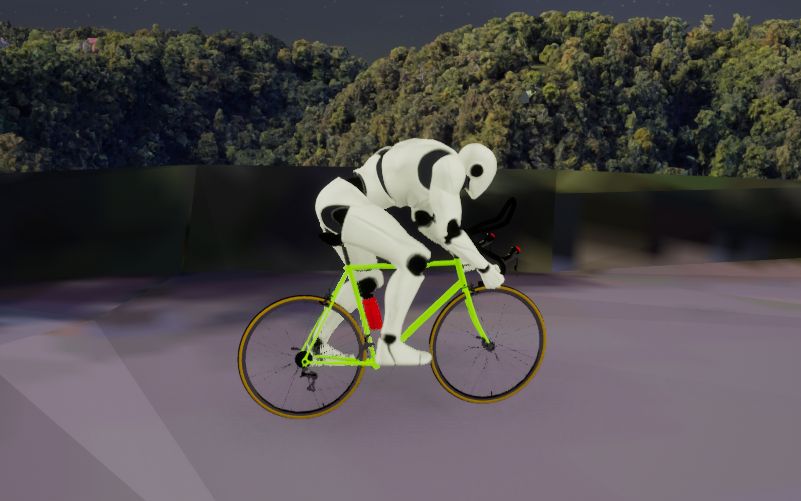}}
	(c)
\end{minipage}
	\begin{minipage}{0.245\linewidth}
	  \centering
  \centerline{\includegraphics[width=\linewidth]{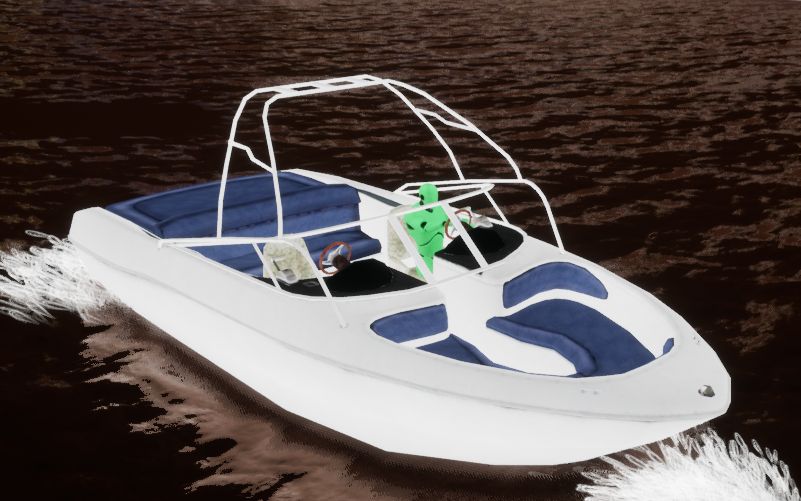}}
	(d)
\end{minipage}
\caption[Pre-defined foreground objects]{Pre-defined foreground objects: (a) drone (b) car (c) cyclist (d) boat.}
  \label{fig:objects}
\end{figure}

As well as the drone object, a further three foreground objects, car, cyclist and boat, have also been integrated into the software tool. Figure \ref{fig:objects} shows example figures for these four objects.

\subsubsection*{Editing Mode}

 \begin{figure}[ht]
\centering
	\begin{minipage}{0.485\linewidth}
	  \centering
  \centerline{\includegraphics[width=\linewidth]{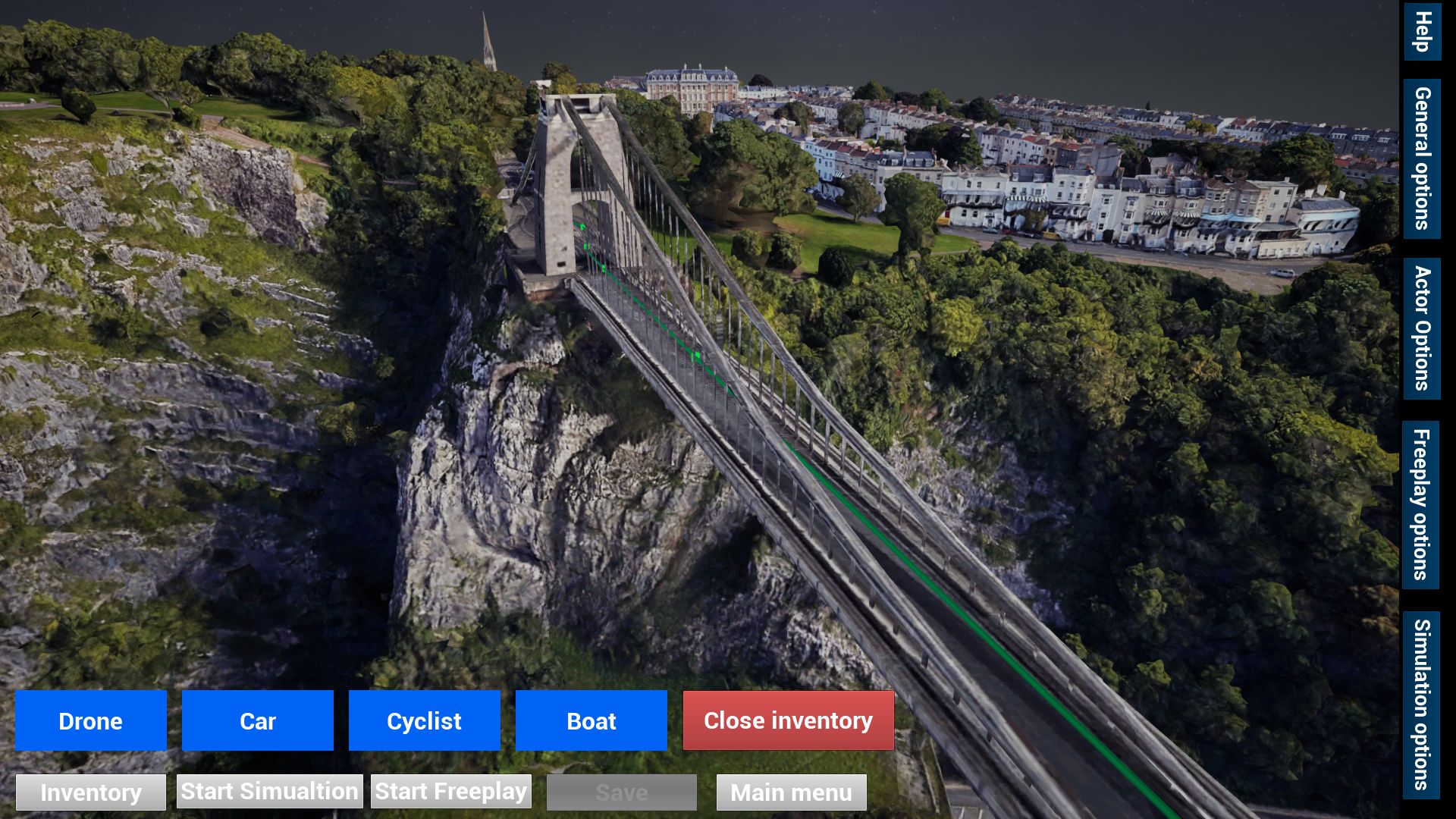}}
	(a)
\end{minipage}
	\begin{minipage}{0.485\linewidth}
	  \centering
  \centerline{\includegraphics[width=\linewidth]{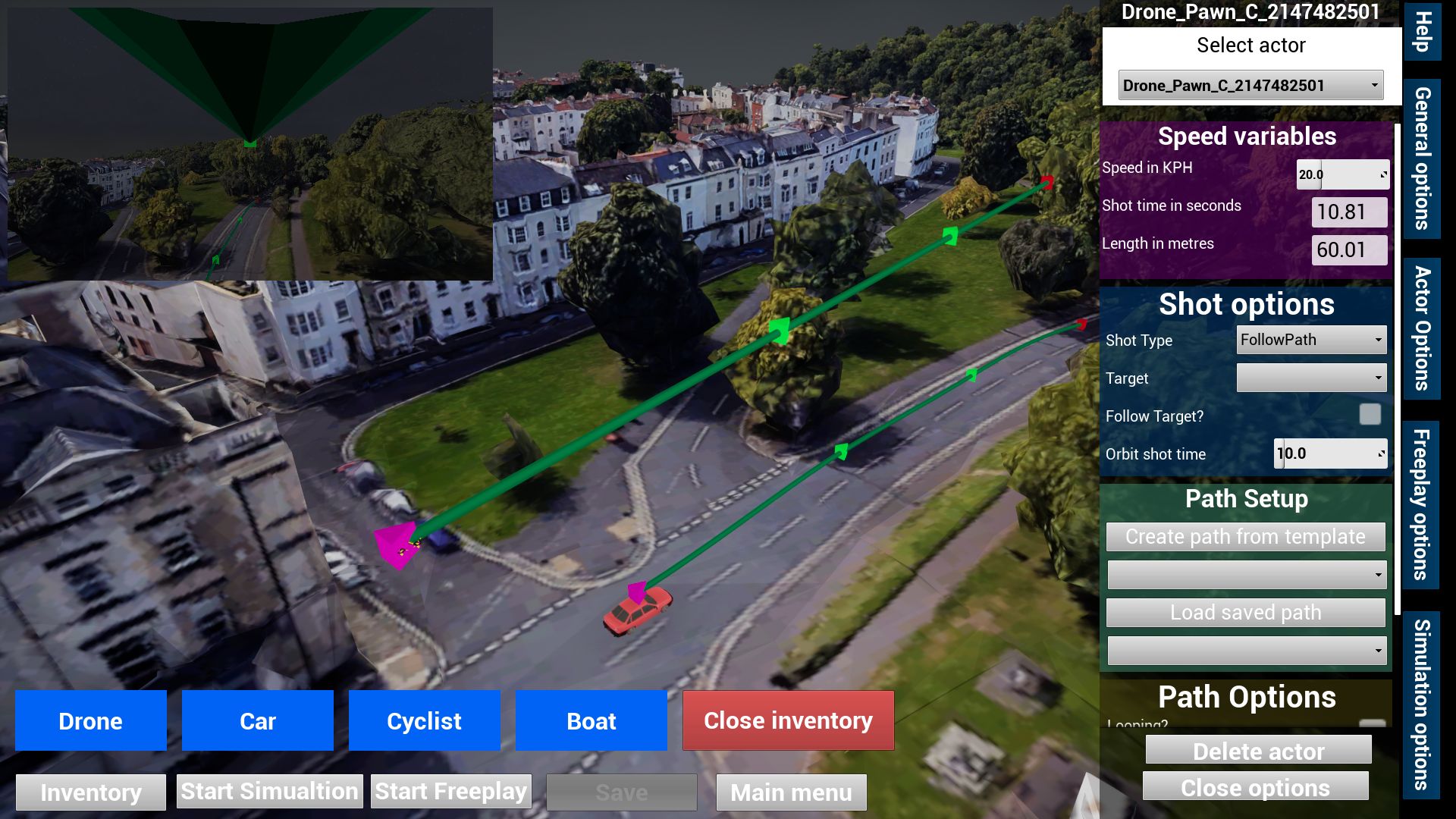}}
	(b)
\end{minipage}
	\begin{minipage}{0.485\linewidth}
	  \centering
  \centerline{\includegraphics[width=\linewidth]{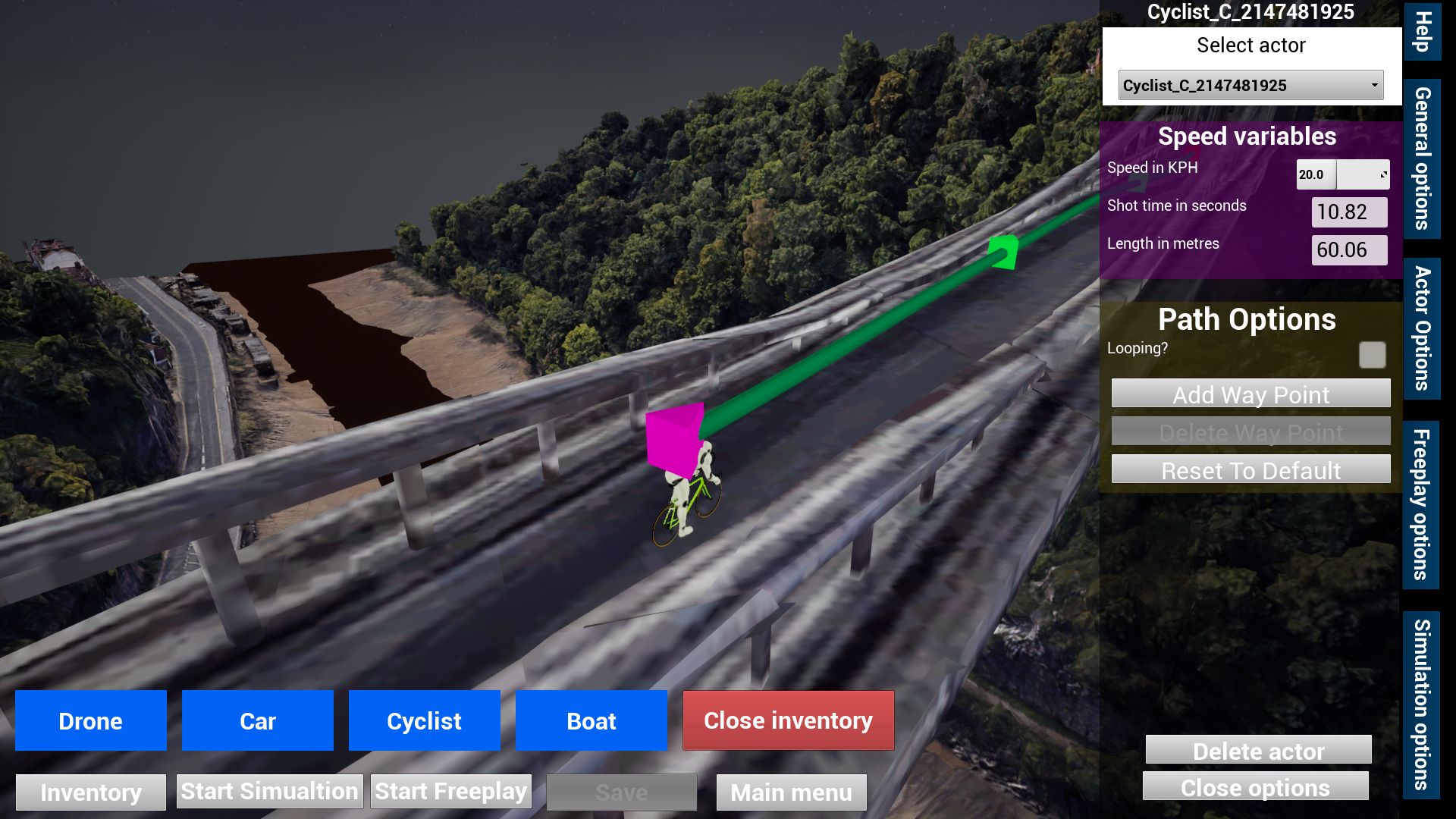}}
	(c)
\end{minipage}
	\begin{minipage}{0.485\linewidth}
	  \centering
  \centerline{\includegraphics[width=\linewidth]{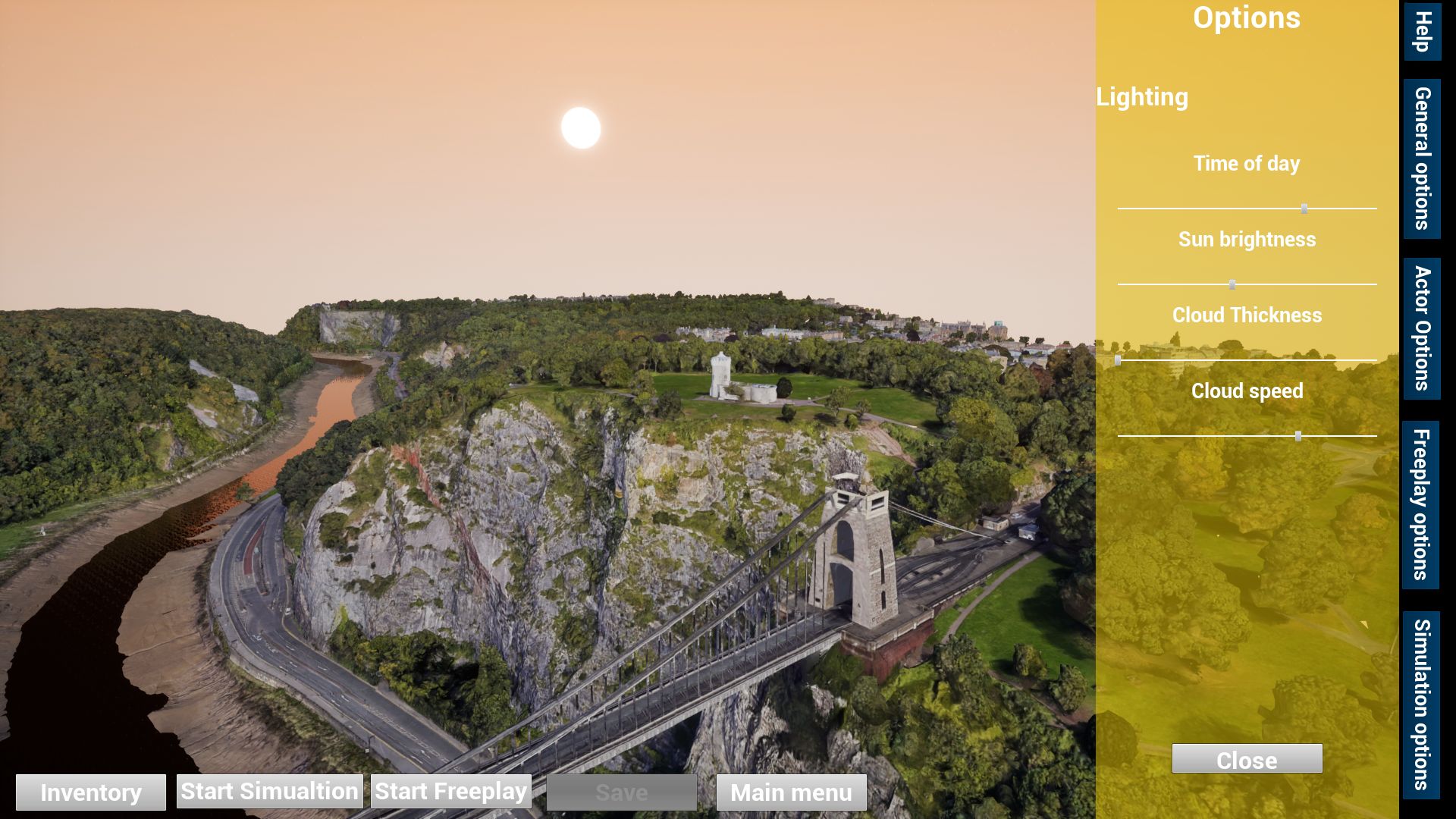}}
	(d)
\end{minipage}
  \caption{(a) The custom interface at the editing mode. (b)The option interface for a drone. (c) The option interface for a cyclist object. (d) The additional options for the map visuals.}
  \label{fig:editing}
\end{figure}


In editing mode, a customised interface has been configured which allows users to drag and drop objects (e.g. drones, cars, cyclists and boats) into the map, as shown in Figure \ref{fig:editing}.(a). Once objects are placed in the map, it is possible to move, rotate and scale the entire object and/or edit its associated waypoints. There are also a range of options for each dynamic object edited via the actor options tab. Figure \ref{fig:editing}.(b) illustrates the options for the drone which include speed, shot type and if the drone will follow the target. Figure \ref{fig:editing}.(c) shows the options for other objects (e.g. car, cyclist and boat), such as speed and path configuration. There are also additional options for the map visuals such as the time of day, lighting, cloud thickness and cloud speed, which are shown in Figure \ref{fig:editing}.(d). 

In order to support simulation of typical shot types in drone cinematography, a range of typical shot trajectories have been configured. Five examples, including ESTABLISH, CHASE, FLYBY, ELEVATOR and ORBIT,  are integrated in the prototype system for the purpose of capability demonstration. All the default parameters recommended are based subjective study results reported in \cite{c:Zhang25} where a camera with 23.66m$\times$13.3mm sensor size and a focal length of 35mm was employed. When different camera setting are selected, these shot parameters will be re-calculated based on the Field of View (FOV) formula show in equation (\ref{eq:fov1}).

\subsubsection*{Simulation Mode}

\begin{figure}[ht]
\centering
	\begin{minipage}{0.485\linewidth}
	  \centering
  \centerline{\includegraphics[width=\linewidth]{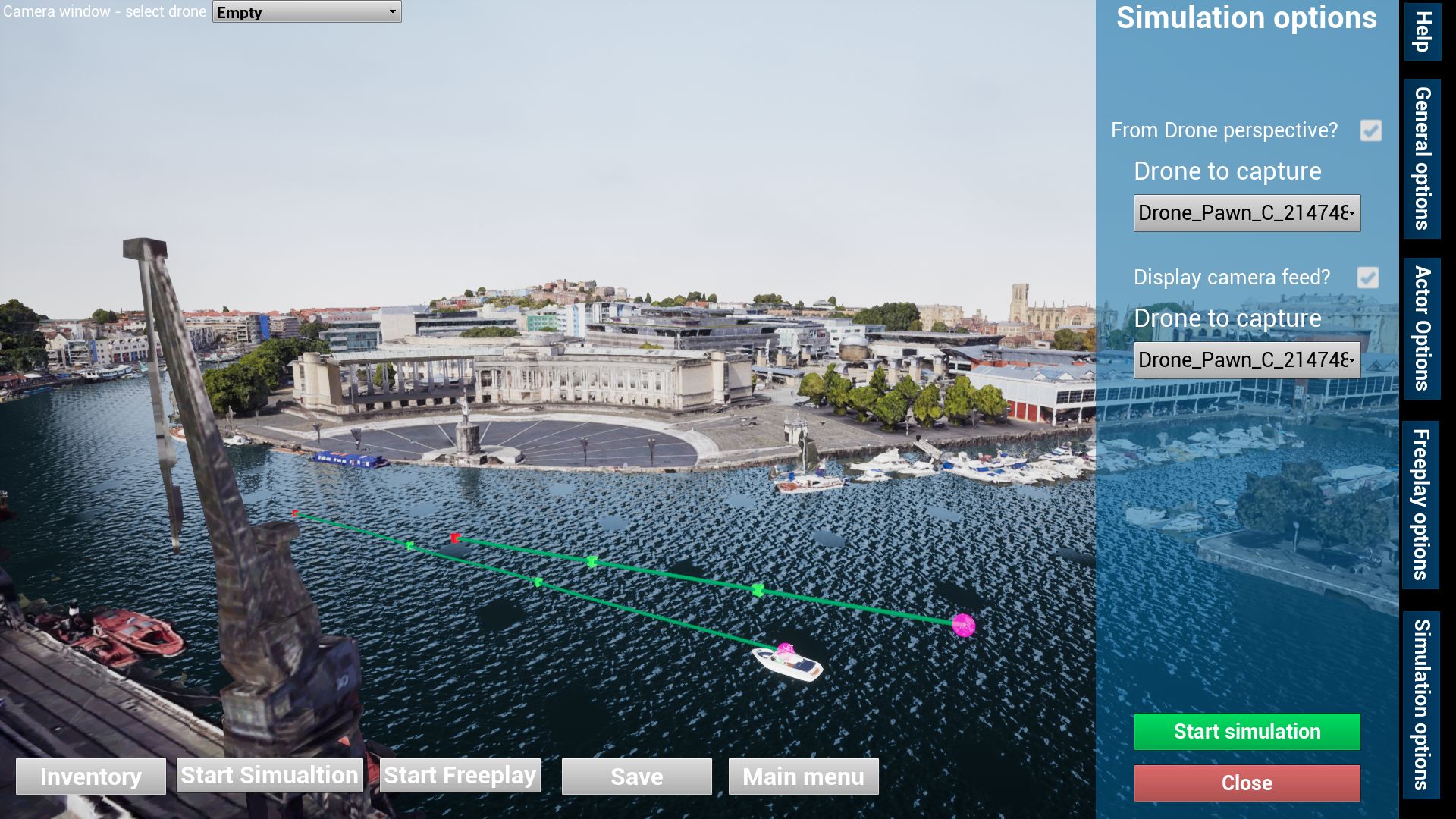}}
	(a)
\end{minipage}
	\begin{minipage}{0.485\linewidth}
	  \centering
  \centerline{\includegraphics[width=\linewidth]{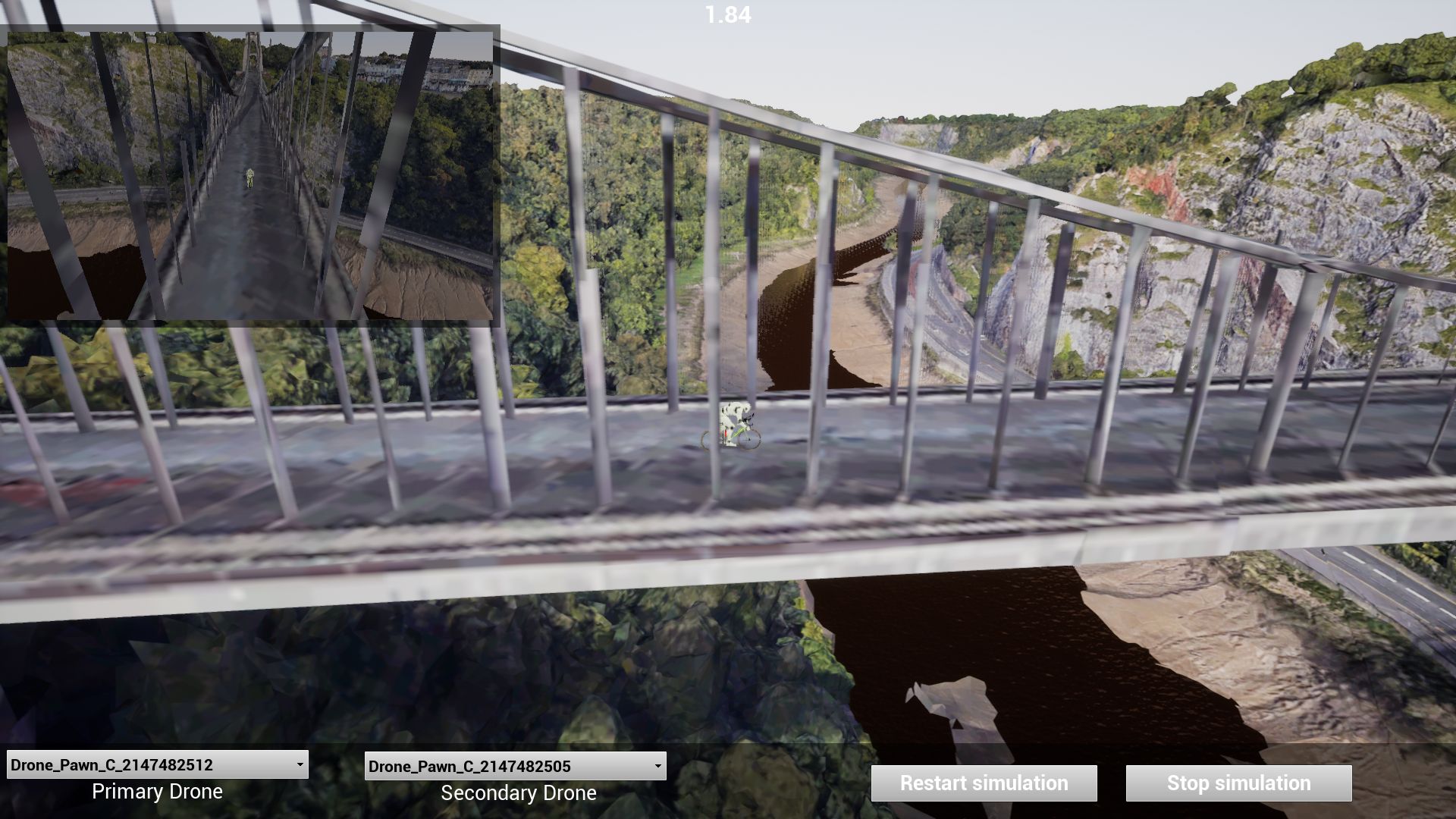}}
	(b)
\end{minipage}
	\begin{minipage}{0.485\linewidth}
	  \centering
  \centerline{\includegraphics[width=\linewidth]{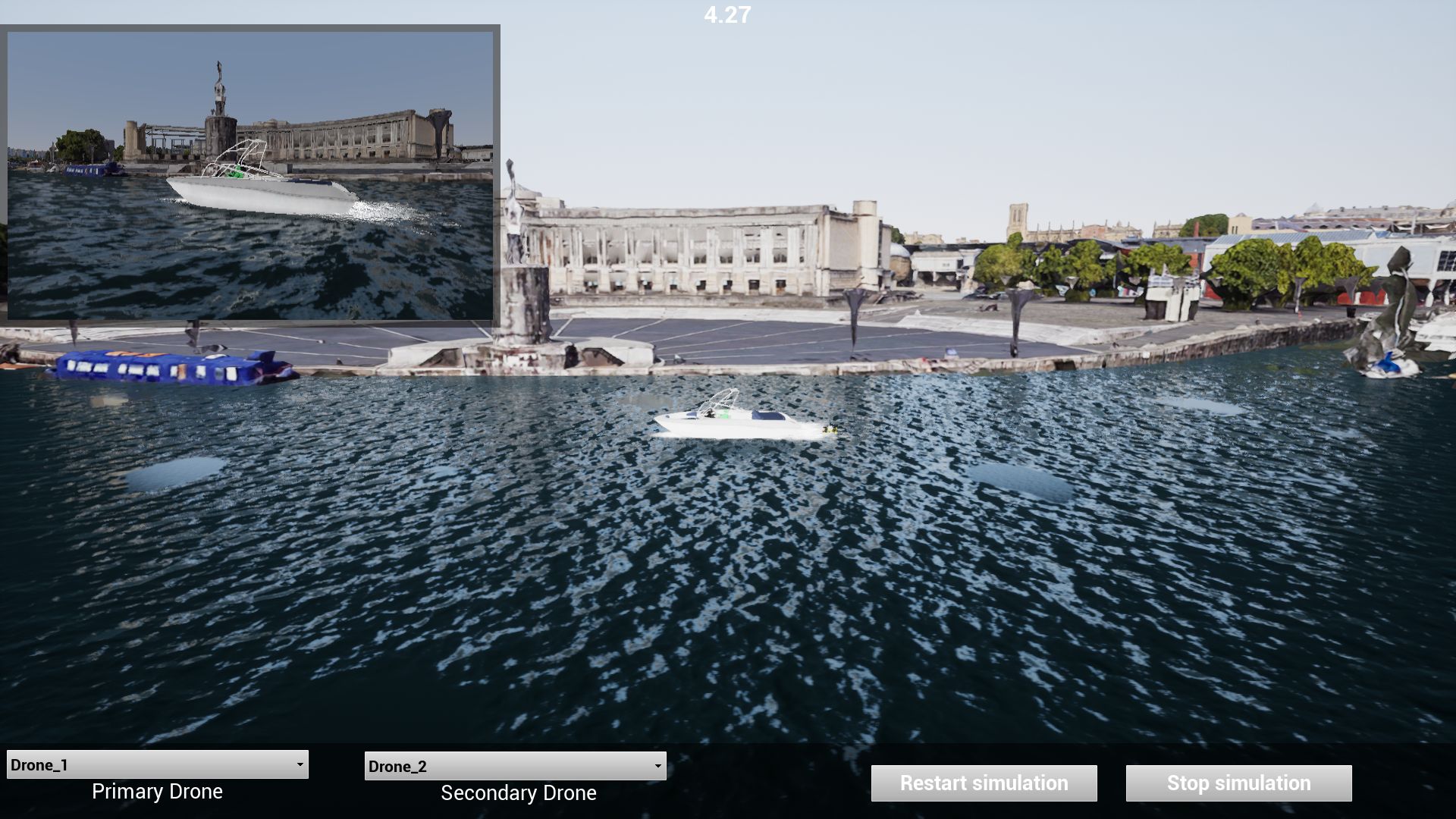}}
	(c)
\end{minipage}
	\begin{minipage}{0.485\linewidth}
	  \centering
  \centerline{\includegraphics[width=\linewidth]{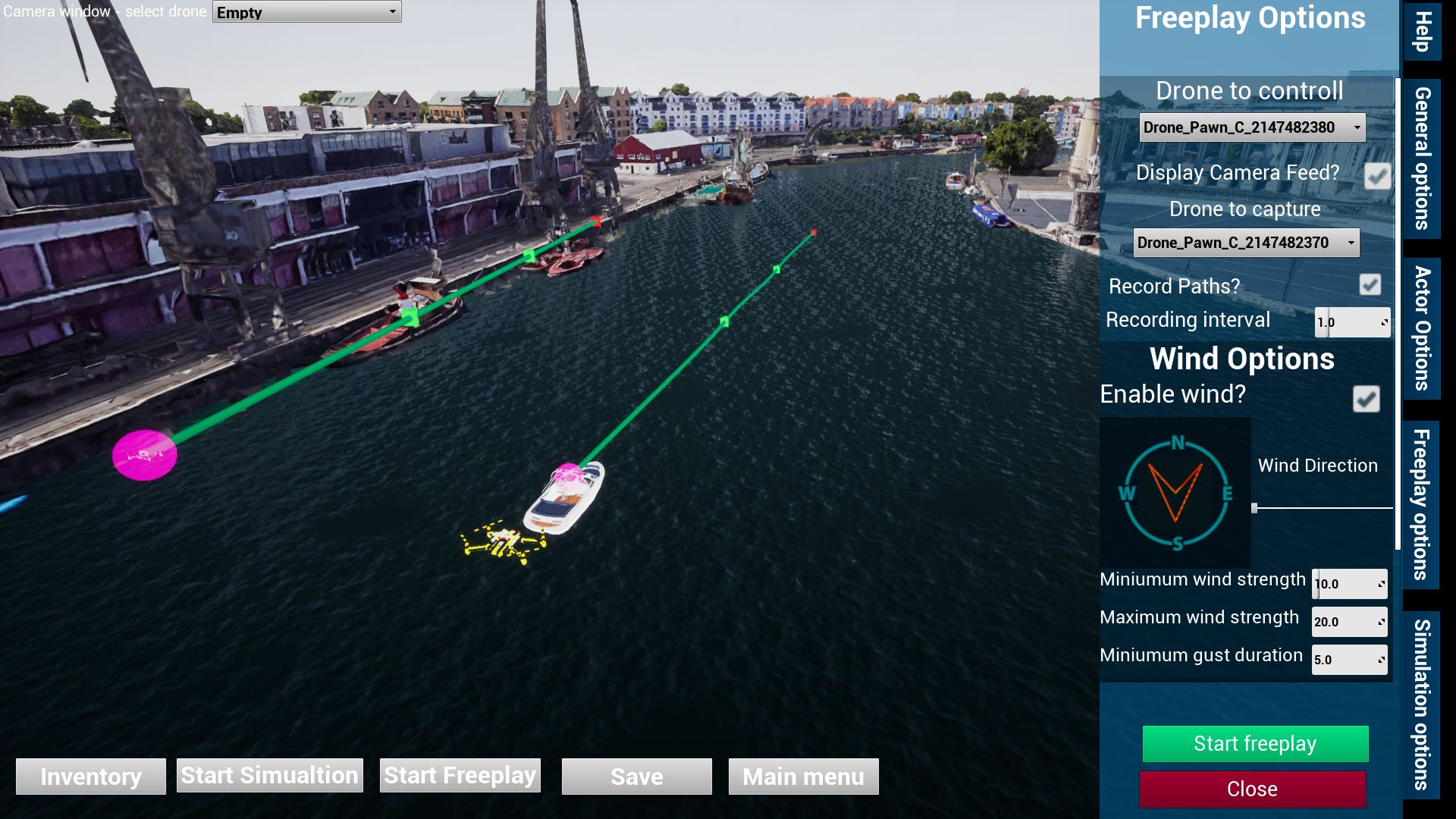}}
	(d)
\end{minipage}
  \caption{(a) The option interface in the simulation mode. (b) A screen shot of simulation mode interface when a cyclist object is moving on a bridge. (c) A screen shot of simulation mode interface when a boat is sailing in a river. (d) The free play option interface.}
  \label{fig:simulation}
\end{figure}

Once the simulation begins, each object within the map will move along the path configured by the user. There is an option to view the simulation either from the floating camera or from a drone perspective. An option is also available to view from the perspective of more than one drone by using the camera window in the top left of the screen. The option interface for the simulation mode is illustrated in Figure \ref{fig:simulation}.(a)-(c).

\subsubsection*{Free Play Mode}

Free-play is similar to the simulation mode except that the user has the option to manually control one of the drones (keyboards and game controllers are currently supported; in future a full drone controller interface will be developed). All the other dynamic objects will behave the same as they would in the simulation mode. There are additional options for the free play mode which include wind parameters and the ability to record the flight path of the manually controlled drone (for off-line evaluation). When the free play mode ends, it is possible to recreate and edit the flight path within editing mode. The option interface for the free play mode is shown in Figure \ref{fig:simulation}.(d).

\subsubsection*{User Feedback and Evaluation}

The current version of this simulation software was demonstrated at the Bristol Drone Cinematography Workshop held in Bristol, UK, in December 2019. Positive (informal) feedback was obtained from all delegates who viewed the demonstration including five international drone cinematography experts. The main challenges remaining are to streamline the environment creation process for users to easily create their own environments, to integrate this software with VR devices to achieve more immersive visual experiences, and to enable its compatibility to primary auto pilot software for autonomous drone operation. Demo videos are available at \url{https://vilab.blogs.bristol.ac.uk/?p=2456} for access.

\subsection*{Conclusion}
\label{sec:conclusion} 

In this paper, a flexible and functional drone training, pre-visualisation and planning simulator is described that is capable of simulation based on actual environments. The tool allows creation of accurate real world environments, and the incorporation of programmable foreground assets as filming targets. It also supports paramaterisable pre-programmed shot types within a user friendly interface which is built upon UE4. Future work will focus on the development of a more flexible environment creation workflow to enable users to create 3D maps for specific areas, and the interface to an HMD to enable a VR platform to achieve more immersive visual experiences.

\subsection*{ACKNOWLEDGEMENTS}

The authors acknowledge funding from the European Union's Horizon 2020 programme (Grant No 731667, MULTIDRONE) and UK EPSRC (The Centre for Doctoral Training In Communications at University of Bristol, EP/I028153/1).

\bibliographystyle{IEEEtran}
\bibliography{IEEEabrv,MyRef}

\end{document}